\documentclass{article}

\PassOptionsToPackage{numbers,sort}{natbib}

\usepackage[preprint]{neurips_2026}

\usepackage[utf8]{inputenc} 
\usepackage[T1]{fontenc}    
\usepackage{hyperref}       
\usepackage{url}            
\usepackage{booktabs}       
\usepackage{amsfonts}       
\usepackage{nicefrac}       
\usepackage{microtype}      
\usepackage{xcolor}         

\usepackage{algorithm}
\usepackage[noend]{algpseudocode}
\usepackage{amsfonts}
\usepackage{amsmath}
\usepackage{amssymb}
\usepackage{array, booktabs, makecell}
\usepackage{babel,blindtext}
\usepackage{graphicx}
\usepackage{subcaption}
\usepackage{xspace}
\usepackage{caption}
\usepackage{xcolor}
\usepackage{listings}
\usepackage{tcolorbox}
\usepackage{wrapfig}

\tcbuselibrary{breakable,skins,listings}

\newtcblisting{promptbox}{
    colback=gray!8,
    colframe=gray!50,
    breakable,
    enhanced,
    left=4pt,
    right=4pt,
    top=2pt,
    bottom=2pt,
    boxrule=0.5pt,
    listing only,
    listing options={
        basicstyle=\small\ttfamily,
        breaklines=true,
        breakatwhitespace=false,
        breakindent=0pt,
        columns=fullflexible,
        keepspaces=true,
    },
}

\newcommand{\algacro}{I-Perceive\xspace}
\newcommand{\projecturl}{\href{https://roboticsjtu.github.io/I-Perceive-Page/}{Project website}}

\newcommand{\appref}[1]{Appendix~\ref{#1}}
\renewcommand{\eqref}[1]{Eq. (\ref{#1})}
\newcommand{\figref}[1]{Fig.~\ref{#1}}

\newcommand{\tabref}[1]{Table~\ref{#1}}

\newcommand{\eg}{\textit{e.g.}}

\title{\algacro: A Foundation Model for \\Vision-Language Active Perception}

\author{%
  Yongxi Huang\textsuperscript{1} \quad
  Zhuohang Wang\textsuperscript{2,3} \quad
  Wenjing Tang\textsuperscript{1} \quad
  Xinyu He\textsuperscript{2} \\
  Cewu Lu\textsuperscript{1,2}\thanks{Corresponding authors.} \quad
  Panpan Cai\textsuperscript{1,2}\footnotemark[1] \\[0.5em]
  \textsuperscript{1}Shanghai Jiao Tong University \quad
  \textsuperscript{2}Shanghai Innovation Institute \quad
  \textsuperscript{3}Beihang University
}

\begin{document}

\maketitle

\begin{abstract}

Active perception—the ability of a robot to proactively select viewpoints to acquire task-relevant information—is essential for robust operation in real-world environments. However, existing approaches are typically limited to fixed objectives or constrained settings, and struggle to generalize to open-ended perception intents specified in natural language.
We propose I-Perceive, a foundation model for language-conditioned active perception in large-scale indoor environments. Given a query image, a set of context images, and a natural language instruction, I-Perceive predicts a 6D camera pose that fulfills the specified perception intent. The model integrates a vision-language pathway for semantic grounding with a geometric reasoning pathway for multi-view 3D understanding, connected via multi-layer semantic fusion to enable language-conditioned geometric reasoning.
To support scalable training, we construct a large-scale dataset of language-viewpoint pairs from both real-world scene-scanning data and simulated environments using an automated pipeline. 
Extensive experiments demonstrate that I-Perceive significantly outperforms strong baselines on prediction accuracy, viewpoint feasibility, and instructions alignment. The model exhibits strong zero-shot generalization to unseen scenes and instructions, and enables closed-loop active perception, progressively refining viewpoints over sequential interactions. \projecturl

\end{abstract}

\section{Introduction}
\label{sec:intro}

Active perception refers to a robot’s ability to proactively control its camera pose in order to acquire task-relevant information for downstream decision-making and execution~\cite{1988ActivePerception, 2018Revisiting, 1988ActiveVision, 2020ViewPlanningSurvey}.
Existing approaches, however, are largely fragmented. On the one hand, many methods are designed for fixed downstream tasks—such as object detection, pose estimation, or scene reconstruction~\cite{wu20153d, sock2017multi, connolly1985determination, mendoza2020supervised}—where the perception objective is hard-coded into task-specific optimization losses. As a result, models developed for different tasks are isolated, difficult to integrate, and lack generality. On the other hand, a large body of work assumes highly constrained environments, typically limited to table-top or small-scale scenes~\cite{zhu2025active, kerr2025eye, liu2026sapave, zou2026activeglasses}, which restricts their applicability to real-world indoor settings.

In this work, we aim to provide a foundation model for active perception that overcomes these limitations, allowing (1) generalizing to open-ended perception intents specified by natural language instructions, and (2) scaling to large-scale indoor environments suitable for mobile manipulators.
Achieving this goal poses several fundamental challenges. First, the model must infer perception intent from natural language instructions, which are inherently diverse and underspecified. Second, it must recover the geometry of large scenes from sparse visual observations, bridging the gap from image space (2D) to geometry space (3D). Third, it must ground language-specified intents into the visual and geometric representations of the scene, requiring alignment across language, image, and geometry spaces. Finally, the model must transform the geometry of task-relevant regions or objects into actionable camera poses, mapping 3D scene understanding to 6D camera poses.

Recent advances provide partial solutions to these challenges. 
Vision-Language Models (VLMs)~\cite{Qwen3-VL, 2023Gemini, karamcheti2024prismatic} excel at aligning language with visual observations, while geometric foundation models (e.g., VGGT)~\cite{2024DUSt3R, 2024MASt3R, 2025VGGT} enable multi-view 3D reasoning and camera pose estimation. 
However, these capabilities remain largely \textit{disconnected}: existing systems lack a principled way to translate open-ended language instructions into viewpoint decisions, leaving a critical gap between semantic intent understanding and geometric action. 

To bridge this gap, we propose \algacro, a foundation model for language-conditioned active perception, that enables end-to-end viewpoint decision-making in large-scale indoor environments.
Given a query image, a set of context images, and a natural language instruction, \algacro predicts a 6D camera pose that fulfills the specified perception intent. 
At a high level, the model integrates two complementary pathways: a \textit{vision-language pathway} that extracts semantic intent from multimodal inputs, and a \textit{geometric reasoning pathway} that models multi-view spatial structure and camera relationships. 
The key challenge lies in tightly coupling these pathways to enable \textit{language-conditioned geometric reasoning under partial observability}.

To this end, we introduce a unified architecture based on \textit{cross-pathway fusion}. 
We represent viewpoint prediction as reasoning over a latent \textit{target frame}, whose learnable tokens participate in joint multi-view attention with observed frames, enabling the model to infer unseen viewpoints under partial observability. 
Semantic information is injected via multi-layer cross-attention through dedicated \textit{semantic tokens}, allowing language-grounded features from the vision-language pathway to progressively modulate geometric reasoning across views. 
This yields a unified model for viewpoint decision-making that jointly reasons about \textit{what to observe} and \textit{where to observe it} in 3D space.

To support scalable training, we construct a large-scale dataset of language-conditioned active perception tasks from both real-world scene-scanning datasets~\cite{dai2017scannet, lazarow2025cubify} and simulated environments~\cite{khanna2023hssd}. 
This results in 162K samples from real-world data and 70K from simulation, enabling both robust generalization and strong instruction-following capability. 
All tasks are annotated with high-quality camera viewpoints using a scalable and fully automated data generation pipeline.


In our experiments, we evaluate \algacro on both in-distribution simulation tasks and out-of-distribution environments.
Results show that \algacro significantly outperforms strong baselines, including vision-language navigation methods~\cite{cheng2024navila}, and next-best-view strategies~\cite{connolly1985determination, mendoza2020supervised}, and VLM prompting, in terms of viewpoint accuracy, feasibility, and task alignment. 
Moreover, \algacro exhibits strong zero-shot generalization to novel scenes and previously unseen perception intents, while robustly handling complex language instructions, rich visual contexts, and partial observability. The model naturally supports a variable number of context images, with performance improving as additional context is incorporated.
These capabilities enable \algacro to perform sequential active perception: when executed multiple times in a closed-loop manner, it progressively refines camera viewpoints, producing progressively-refined views over successive perception steps.

\begin{figure}[!t]
    \vspace*{-1.2cm}
    \centering
    \includegraphics[width=\linewidth]{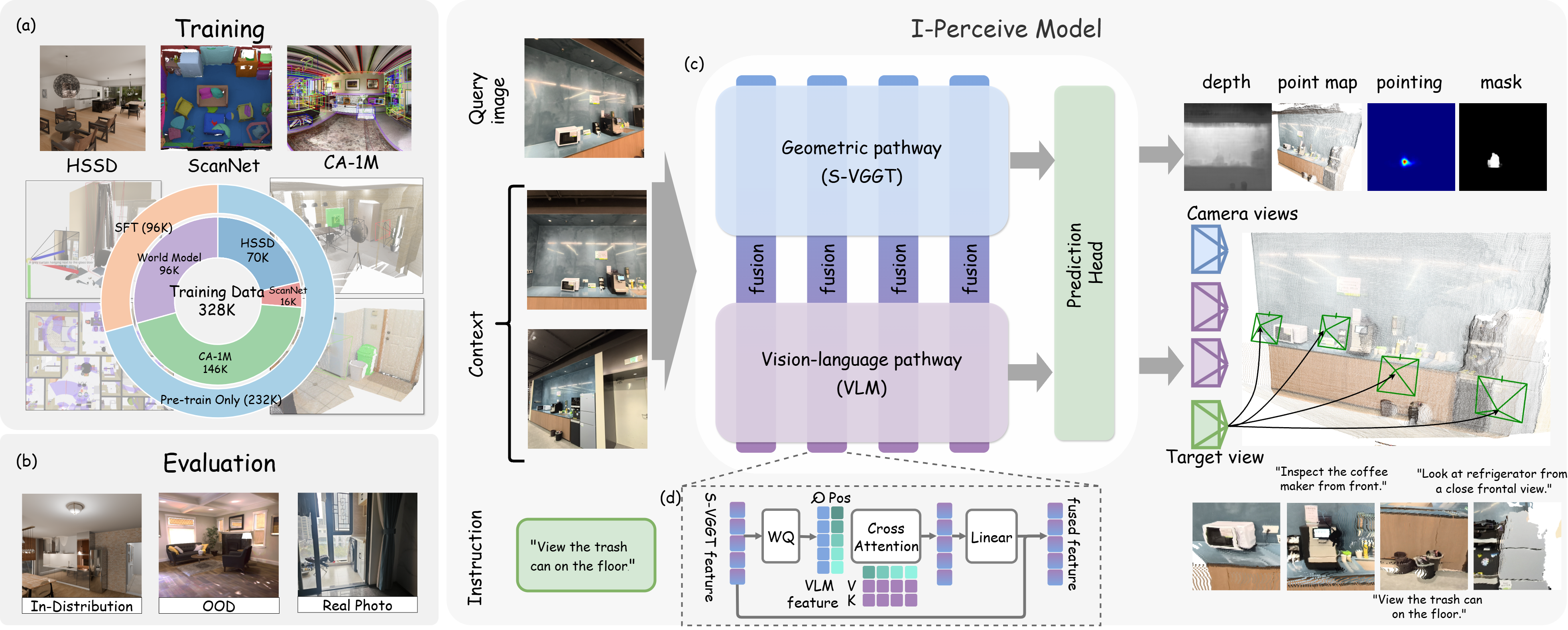}
    \captionof{figure}{\textbf{\algacro} is a foundation model for vision-language active perception. Given context images and a natural language instruction, the model predicts a target camera pose that fulfills the observation intent. By fusing VLM semantic features into a geometric backbone, \algacro demonstrates strong zero-shot generalization to unseen real-world environments.}
    \label{fig:overview}
\end{figure}

\section{Related Work}
\label{sec:related}
\subsection{Active Perception}

Active perception refers to an agent's ability to actively control its sensors to acquire task-relevant information. As formulated by Bajcsy~\cite{1988ActivePerception,2018Revisiting}, an active perceiver must determine \textit{what} to sense, and consequently decide \textit{how} and \textit{where} to perceive. This paradigm has motivated extensive research across robotics and computer vision~\cite{1988ActiveVision,2020ViewPlanningSurvey}.

\textit{Reconstruction-Driven Approaches.}
A major line of active perception research focuses on 3D reconstruction, where the objective is to maximize geometric completeness. Classical Next-Best-View (NBV) methods adopt generate-and-test strategies, selecting viewpoints that reveal the largest amount of unknown geometry~\cite{connolly1985determination,wong1999next}. Subsequent works introduced information-theoretic criteria to quantify viewpoint utility~\cite{2016InfoGain,2018InfoGainComparison}, while learning-based approaches directly predict informative viewpoints from partial observations~\cite{mendoza2020supervised,peralta2020next}. Recent work further formulates reconstruction-oriented active perception as informative path planning under partial observability~\cite{qu2026efficient}. However, these methods primarily optimize geometric coverage and lack the semantic understanding required to interpret diverse user intents.

\textit{Perception-Driven Approaches.}
Active perception has also been studied for object recognition and pose estimation tasks. Early work integrated sensor planning with object hypothesis generation~\cite{hutchinson1988planning}, while deep learning approaches such as 3D ShapeNets~\cite{wu20153d} select viewpoints that minimize classification uncertainty. Other methods leverage multi-view reasoning for recognition~\cite{johns2016pairwise} and 6D pose estimation in cluttered scenes~\cite{sock2017multi,2024ActPerMoMa}. Recent work~\cite{liu2026sapave,zou2026activeglasses} further integrates active perception with manipulation by jointly learning camera motion and manipulation policies, but remains largely restricted to table-top environments.
For large-scale environments, Vision-Language Navigation (VLN)~\cite{anderson2018vision,2024VLNSurvey,hong2021vln} is also closely related, where agents navigate toward language-specified goals. Recent VLN methods leverage foundation models for semantic grounding and reasoning~\cite{khandelwal2022simple,majumdar2022zson,zhang2024navid,2024vlfm}. However, although VLN implicitly involves viewpoint selection, its objective remains navigation-oriented, where views are primarily chosen based on exploration or proximity to goal locations rather than perception quality.

\textit{Positioning of Our Work.}
Existing active perception methods are typically designed for fixed objectives (\eg, reconstruction, recognition, or navigation) with task-specific optimization criteria. In contrast, our work studies \textit{instruction-driven} active perception that generalizes across diverse perception intents specified in natural language. This requires jointly reasoning about semantic intent, visual observations, and 3D geometry under partial observability, a capability that remains largely absent in prior work.

\subsection{Vision-Language and Geometry Foundation Models}

\textit{VLMs and Spatial Understanding.}
Vision-Language Models (VLMs) have achieved remarkable success in semantic understanding and reasoning across vision and language modalities
~\cite{Qwen3-VL, 2023Gemini, karamcheti2024prismatic}.
However, recent studies reveal that VLMs struggle with spatial reasoning and 3D geometric understanding~\cite{yang2025thinking, zhang2025mllms}. Some works apply VLMs directly to active perception~\cite{zhu2025active, sripada2024ap,kerr2025eye}, but these typically operate in constrained settings (\eg, table-top scenes) with limited action spaces (\eg, image cropping or zooming), lacking fundamental 3D scene understanding.

\textit{Geometric Foundation Models.}
Recently, transformer based geometric foundation models ~\cite{2024DUSt3R,2024MASt3R,2025VGGT} have shown impressive capabilities in recovering 3D structure and camera poses from sparse views. By leveraging large-scale training data and powerful attention mechanisms, VGGT~\cite{2025VGGT} achieves robust spatial reasoning across diverse scenes and viewpoints, enabling effective mapping from discrete images to holistic 3D scene understanding and camera pose inference. 
However, while VGGT-like models are strong at estimating camera poses from given observations, they are not designed to perform viewpoint decision-making under partial observability, and do not accept language input.


\textit{Bridging Semantics and Geometry.}
Recent work has explored integrating VLM semantics with geometric reasoning. 
One line of research enhances VLMs' spatial understanding through language-level supervision from geometry models ~\cite{chen2024spatialvlm, zhou2025robotracer, cao2025spatialdreamer}, while another incorporates explicit geometric features from geometric foundation models into VLMs to improve downstream tasks such as spatial QA and manipulation action generation ~\cite{lin2025evo}. 
These approaches primarily leverages geometry to enhance VLM-centric reasoning, demonstrating the feasibility of joint semantic-spatial reasoning.

\textit{Positioning of Our Work.} \algacro performs the integration in an \textit{geometry-centric} manner: it injects VLM-derived semantic features into a geometric reasoning backbone as conditions. 
The core insight is that viewpoint decision-making is fundamentally a geometric problem, requiring reasoning in 3D space. Images and language only specify the semantic and task contexts. 

\section{Overview}

\algacro formulates active perception decision-making as a mapping from multi-view scene observations and language tasks to executable camera poses. 
The input include a sequence of RGB images of the environment and a natural language instruction specifying the perceptual intent.
The first image frame is the referred to as the \textit{query frame}, representing the current view of the robot; the subsequent frames serve as \textit{context frames}, providing auxiliary information of the global scene.
A model should output a target camera pose that is both physically feasible (executable by a physical robot) and accurately fulfills the intent of language task.



The core challenge lies in jointly reasoning over language, visual observations, and 3D scene geometry under partial observability.
To address this, \algacro integrates two complementary components: a vision-language pathway for extracting semantic intent from multimodal inputs, and a geometric reasoning pathway for modeling 3D structure and camera pose relationships across views. 
Crucially, semantic features are progressively injected into the geometric pathway via multi-layer cross-attention, enabling \textit{language-conditioned geometric reasoning}.
\algacro is trained on a large-scale dataset of active perception tasks constructed from both real-world scene-scanning data and simulated environments, enabling both robust generalization and strong instruction-following performance.
At inference time, the model supports both single-step prediction and closed-loop execution, where newly observed views are iteratively incorporated as context to refine future predictions. 
\figref{fig:overview} provides an overview of the method. The accompanying video provides a detailed narrative.

\section{Architecture of the \algacro Model }
\label{sec:model}

\begin{figure*}[t]
  \centering
  \begin{minipage}[t]{0.49\textwidth}
      \vspace{0pt}
      \centering
      \includegraphics[width=\linewidth]{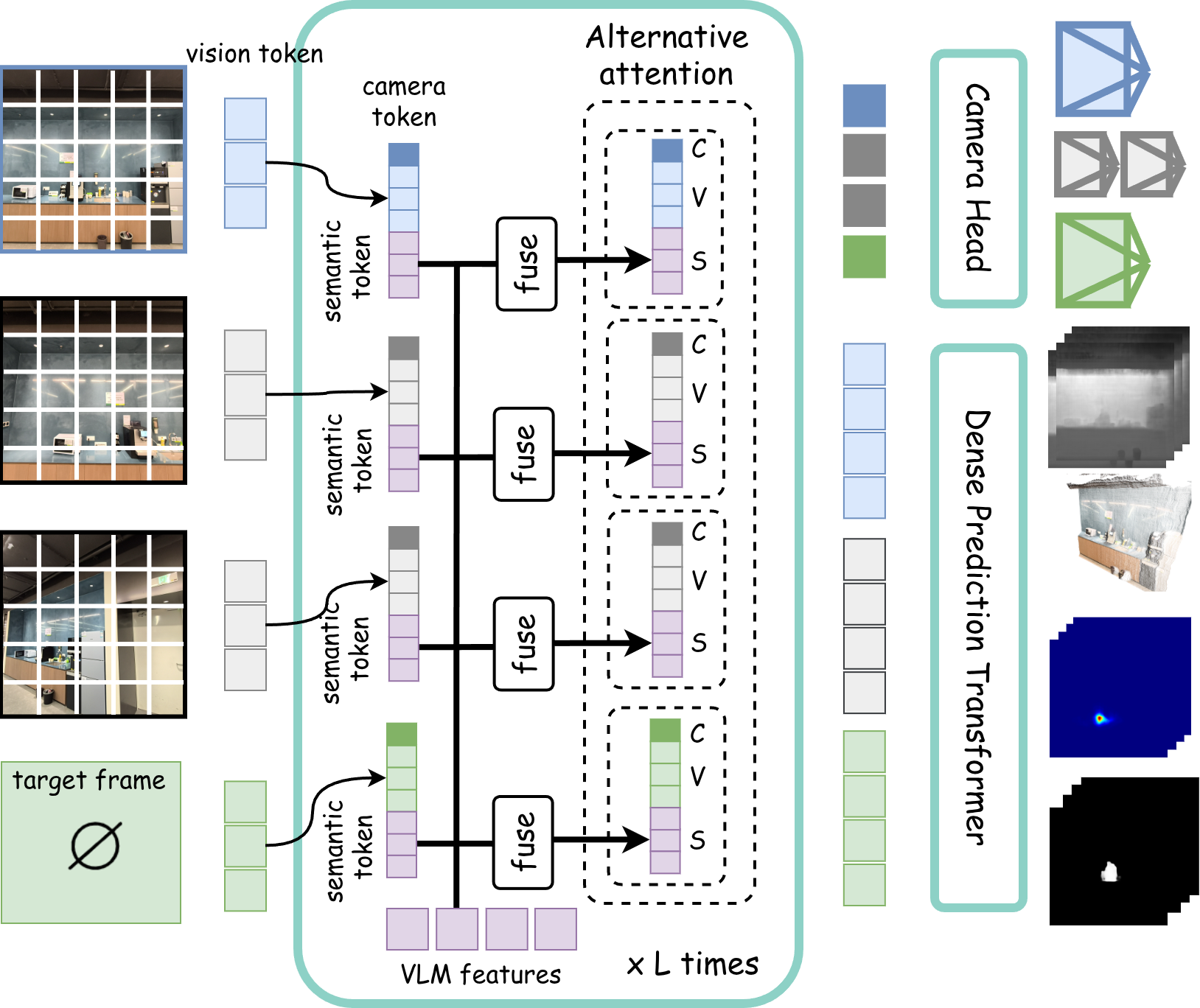}
  \end{minipage}
  \hfill
  \begin{minipage}[t]{0.49\textwidth}
      \vspace{0pt}
      \centering
      \includegraphics[width=\linewidth]{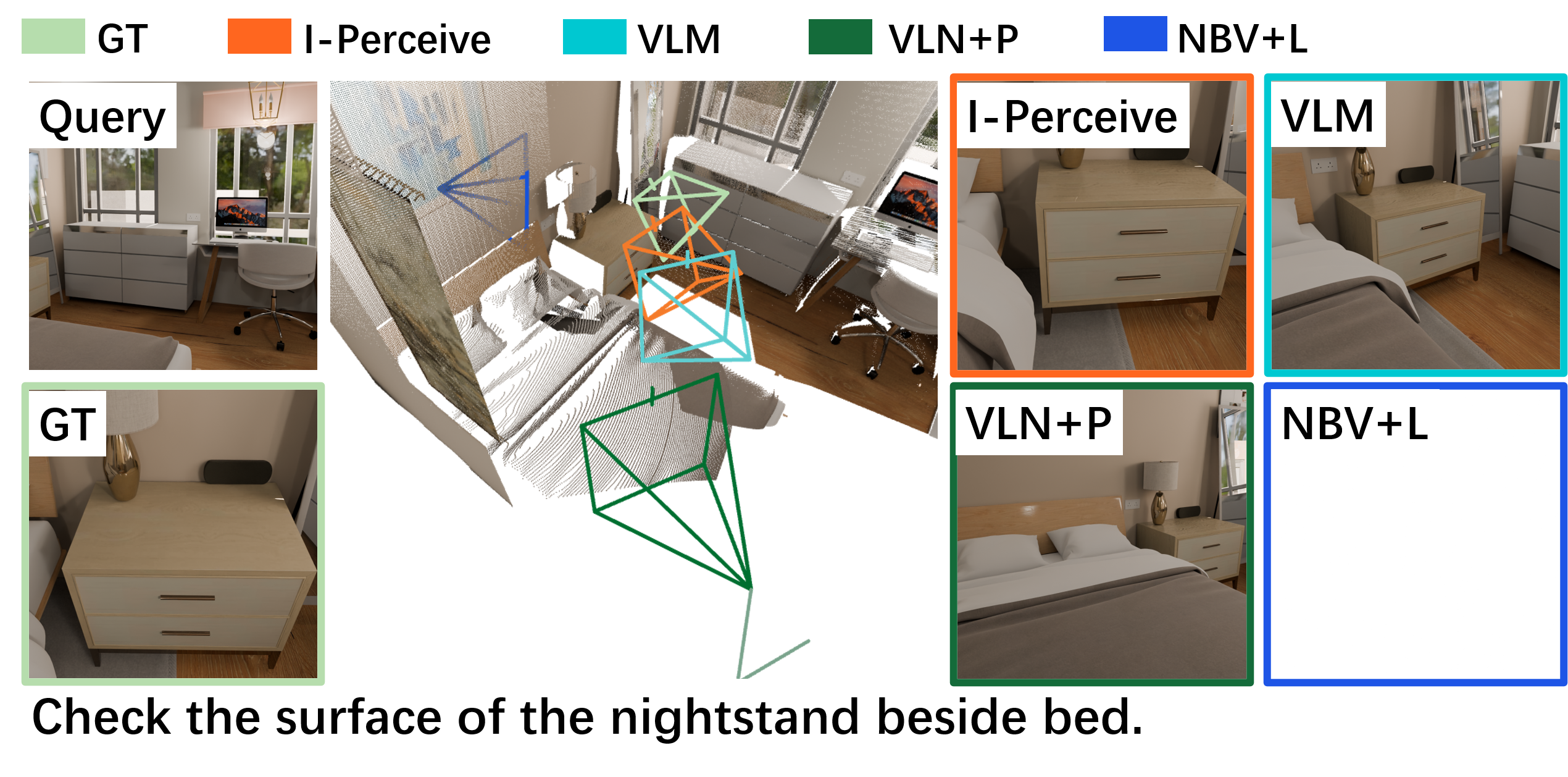}\par\vspace{2pt}
      \includegraphics[width=\linewidth]{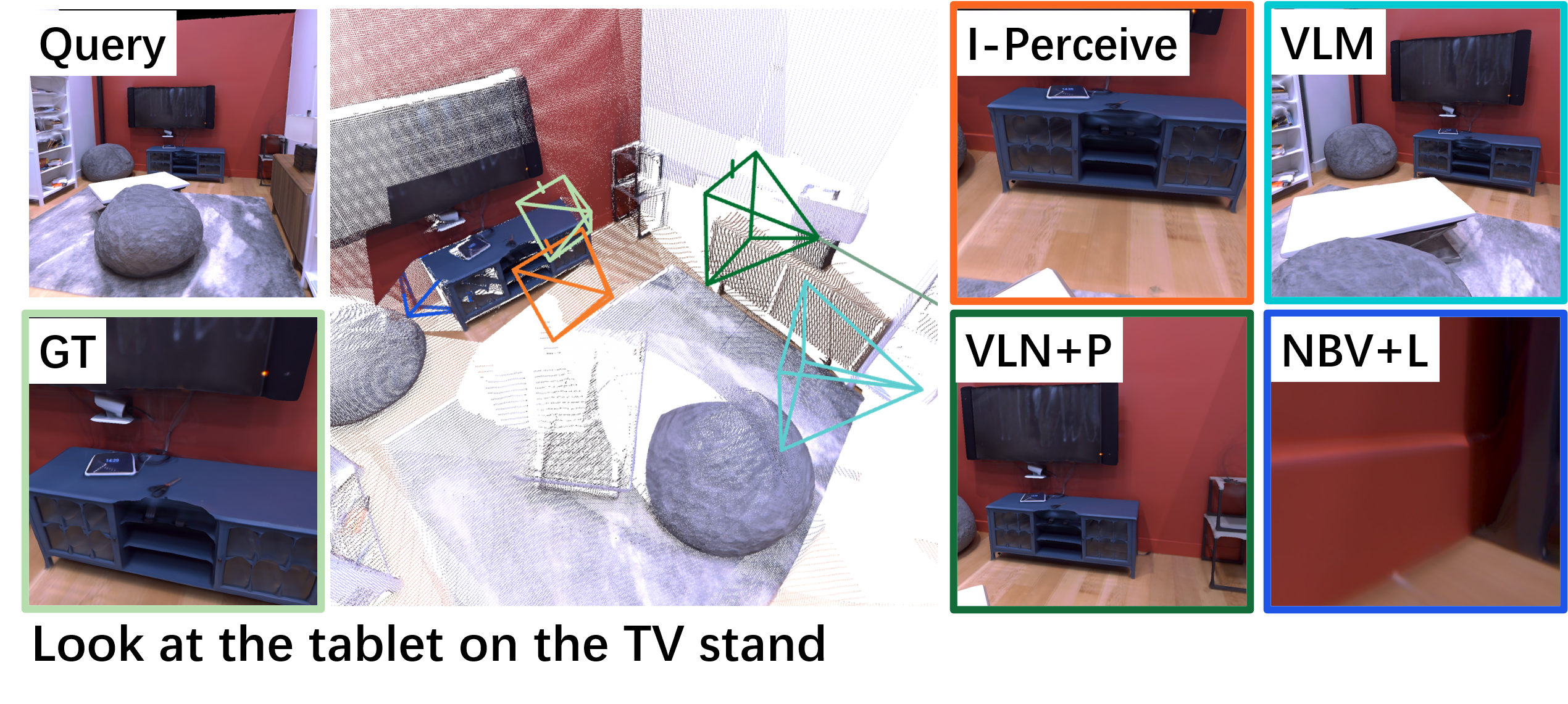}
  \end{minipage}

  \caption{
  \textbf{Left:} A zoom-in view of \algacro's geometric-reasoning pathway.
  \textbf{Right:} Qualitative results of \algacro.
  }
  \label{fig:arch_result}
  \label{fig:geometric_pathway}
  \vspace{-10pt}
\end{figure*}



\algacro maps visual context and language instructions to a target camera pose in an end-to-end manner, performing joint semantic and geometric reasoning using two tightly-integrated pathways. 
The detailed architecture is as follows.

\paragraph{Input and Output.}
The model takes as input $N$ RGB frames $\{I_i\}_{i=0}^{N-1}$ and a natural language instruction $\ell$.
The goal is to predict the target camera parameters $\mathbf{g}_N$ that fulfill the instruction, defined w.r.t. the coordinate system rooted at the camera pose of $I_0$.
Each camera is parameterized as $\mathbf{g}_i = [\mathbf{q}_i, \mathbf{t}_i, \mathbf{f}_i] \in \mathbb{R}^9$, representing the rotation quaternion, translation vector, and field of view, respectively.
In addition to camera pose prediction, the model optionally produces dense geometric outputs (depth and point maps) and semantic masks (2D pointing and target object masks), comprising auxiliary tasks to facilitate spatial reasoning and semantic grounding during training.

\paragraph{Vision-Language Pathway.}
We use a pretrained VLM (Qwen3-VL-4B) backbone to encode multimodal semantic information from images and language. To extract semantic representations, instead of using the final-layer features, we extract a rich set of intermediate key-value (KV) features from selected transformer layers $\mathcal{L}=\{l_j\}_{j=1}^{N_Q}$:
\begin{equation}
    K^{(l_j)}, V^{(l_j)} = \text{ExtractKV}(\text{VLM}, l_j, \mathcal{I}, \ell)
    \label{eqn:vlm_features}
\end{equation}
These multi-layer representations reflect the progressive semantic reasoning structure, allowing smooth integration to progressive geometric reasoning. 




\paragraph{Geometric Reasoning Pathway (\figref{fig:geometric_pathway}).}
We adopt a multi-view transformer architecture for geometric reasoning, 
inspired by a prior geometric foundation model, VGGT~\cite{2025VGGT}, that recover 3D structure and camera poses from visual observations. 
However, unlike VGGT, which focus on reconstructing geometry and estimating poses of observed views, our goal is to infer unseen, task-driven viewpoints conditioned on language and partial observations.

Similar to \cite{2025VGGT}, the geometric backbone of \algacro maps each input frame as a set of tokens encoding visual appearance (\textit{vision tokens, $v$}) and camera state (\textit{camera tokens}, $c$), using a stack of attention layers that alternate intra- and inter-frame attentions, to perform joint reasoning across all frames and model their spatial relationships. 
This enables the model to aggregate multi-view information and build an implicit 3D understanding of the scene.

To enable language-conditioned active perception, we uniquely introduce two key design elements:

\textit{(1) Target-frame tokens.}
We append a latent target frame $I_T$ with learnable tokens to represent an unseen viewpoint.
Its vision and camera tokens participate in the same alternating attention pathway with the start and context frames, allowing to aggregate information from the observed frames.
This effectively casts viewpoint decision-making as extrapolation from seen to unseen viewpoints.

\textit{(2) Semantic tokens.}
We introduce \textit{semantic tokens} to each frame, denoted as $s$, that serve as carriers of language-grounded information.
In intra-frame attention, semantic tokens interact with both vision tokens $(v)$ and VLM features via \eqref{eqn:fusion}, enabling local grounding of semantics in visual content.
These fused representations are then propagated globally via inter-frame attention.

\paragraph{Cross-Pathway Fusion.}
Language-conditioned information are repeatedly injected into semantic tokens, via cross-attention:
\begin{equation}
    s_{i}' = s_{i} + \gamma \cdot \text{Proj}\left(\text{Attn}\left(W_Q s_{i}, K^{(l)}, V^{(l)}\right)\right), \forall l \in \mathcal{L}.
    \label{eqn:fusion}
\end{equation}
Here, $\mathcal{L}$ denote layers where we extract VLM features, $K^{(l)}$ and $V^{(l)}$ represent the extracted features, respectively.
Compared to shallow or late fusion strategies, injecting semantic information at multiple layers enables deep integration of language and geometry, allowing semantic intent to progressively modulate spatial reasoning throughout the network.
See more implementation details in \appref{app:model_details}.

\paragraph{Prediction Heads.}
The final token representations are decoded to predict camera poses, dense, and semantic output for both observed frames and the target frame.
Concretely, camera tokens at selected layers are aggregated and passed through a regression head to estimate camera parameters of all frames; vision tokens at selected layers are aggregated and fed to transformer-based dense prediction heads for depth map and point map generation, in order to enhance 3D structural reasoning.
Vision tokens are also fed to semantic grounding heads, to generate 2D pointing heatmaps and target-object masks, in order to encourage semantic understanding.
At inference time, only the camera pose prediction head is required; other predictions are optional.

\section{Training \algacro}
\label{sec:formulation}

We train \algacro using supervised learning. 
Importantly, there exists no public dataset that directly supports language-conditioned active perception learning, which requires pairing multi-view observations and language tasks with camera-view annotations.
To address this, we construct a large-scale \algacro dataset of vision-language-viewpoint tuples, through an automated data generation and labeling pipeline, which transforms indoor scene datasets into active perception tasks with minimal human intervention.
Each sample in the dataset consists of a sequence of $N+1$ frames $(\mathcal{F}_i)_{i=0}^{N}$ and a natural language instruction $\ell$.
The first $N$ frames $\{\mathcal{F}_i\}_{i=0}^{N-1}$ are observed views, and the last frame $\mathcal{F}_N$ is the target view that satisfies the intent.
Each frame $\mathcal{F}_i = (I_i, D_i, M_i, \mathbf{g}_i)$ contains an RGB image $I_i \in \mathbb{R}^{3 \times H \times W}$, a depth map $D_i$, a semantic mask $M_i$, and camera parameters $\mathbf{g}_i$. 
\subsection{The \algacro Dataset}
\label{sec:dataset}

We leverage two complementary sources of data: real-world scene-scanning datasets, including ScanNet~\cite{dai2017scannet} and CA-1M~\cite{lazarow2025cubify}, and simulated indoor scenes from HSSD~\cite{khanna2023hssd}. We use the following pipeline to construct active perception tasks:


\noindent \textit{(a) Target sampling.}
For each data point, we first sample a target frame $\mathcal{F}_N$, and the camera pose $\mathbf{g}_N$, corresponding to meaningful observation intents using object-centric heuristics, leveraging dataset-specific annotations (e.g., masks, bounding boxes, or asset templates).
For ScanNet and CA-1M, target frames are sampled from the scanning video, by maximizing target visibility and centered framing; for HSSD, we construct reusable target-pose templates for each object type, with predefined sampling ranges and perception intents.
Across all datasets, we apply filtering to reject frames with collision or insufficient visibility, producing semantically meaningful and geometrically feasible target views.

\noindent \textit{(b) Query and context sampling.}
Given a target viewpoint, we then sample the query and context frames, $\{\mathcal{F}_i\}_{i=0}^{N-1}$, together with the camera poses $\{\mathbf{g}_i\}_{i=0}^{N-1}$, to emulate partial scene observation.
For real-world scanning datasets, since images are already collected along scanning trajectories, we sample query and context frames around the target while ensuring consecutive camera-pose proximity and the overall view diversity.
For HSSD, we explicitly plan collision-free paths among sampled target poses, and sample query and context frames along the resulting trajectories.
This trajectory-based sampling better mimics online robot perception, where context images are accumulated sequentially.


\noindent \textit{(c) Free-form instruction generation.}
We use a VLM to generate natural language instructions $\ell$ conditioned on the target view image, verbalized viewpoint attributes (e.g., relative offset, viewing direction, and framing), and semantic description of the target object when available.
The model converts these inputs into free-form instructions that specify what to observe and how to view it, covering diverse intent types such as object-centric, spatial, viewpoint-specific, and functional descriptions.
Examples and prompts are provided in \appref{app:instruction_generation}.

\noindent \textit{(d) Disambiguation with 2D pointing.}
When an instruction refers to multiple similar objects or regions, we use a VLM to detect ambiguity and augment the instruction $\ell$ with a 2D point reference (``\textit{(x, y) in image}'') to disambiguate the intended target.

\noindent \textit{(e) World-model-based view improvement.}
We refine target viewpoints using a generative world model to increase diversity and naturalness.
A VLM first converts the instruction of the active perception task into a camera-motion prompt, which is used by a video generation model (Wan2.2\cite{wan2025}) to synthesize a intent-consistent video.
We then use VGGT to extract the precise camera motion from the video, and apply it to adjust the target camera pose $\mathbf{g}_N$. 
This improves the realism of target views and their alignment with task instructions.

With the above pipeline, steps (a)--(d) yield a pre-training dataset with 162K samples from ScanNet and CA-1M, and 70K from HSSD. 
Step (e) is applied to $\sim$30\% of CA-1M scenes, producing an additional 96K high-quality samples for post-training.
See additional pipeline details in \appref{app:data}.

\subsection{Training Details}
\label{sec:learning}

We first train \algacro on the pre-training dataset for 5 epochs, with VLM backbone frozen and all learning targets enabled. Then we fine-tune the model with the enhanced post-training dataset for 3 epochs, unfreezing VLM backbone with LoRA~\cite{hu2022lora} and disabling semantic mask and point heatmap losses to focus on instruction following. The training process takes 30 hours on 32 H200 GPUs.
See further details in \appref{app:model_training}.






\section{Experiments}
\label{sec:experiments}

\paragraph{Benchmarks.}
We evaluate \algacro on both in-distribution and out-of-distribution settings across diverse indoor environments.

\textit{(a) HSSD (in-distribution).}
We construct a set of 52 novel tasks with scene-level splits, where all test scenes are entirely unseen during training. To ensure credible and unbiased evaluation, target viewpoints are annotated by human experts rather than generated by the data pipeline.

\textit{(b) Replica (out-of-distribution).}
To assess zero-shot generalization, we construct a new test set on the Replica scene dataset~\cite{straub2019replica}, which provides reconstructed real-world indoor environments. We generate 89 novel language-conditioned tasks, with target views annotated by human experts.

\paragraph{Baselines.}
We compare \algacro against a range of relevant baselines. See detailed implementations of baselines in \appref{app:baselines}.
All VLM components used by the baselines are GPT-5.4.

\noindent\textit{(a) Language-Conditioned Next-Best-View (NBV+L).}
NBV methods are typically designed for coverage maximization and do not directly accept language tasks. 
To enable comparison, we implement a language-conditioned variant that samples candidate viewpoints around the target region and uses a VLM to select the best view based on instruction alignment. 

\noindent\textit{(b) Vision-Language Navigation with Perception View Selection (VLN+P).}
We then implement a VLN-style baseline. It first navigates toward the target object based on language instructions using NaVILA~\cite{cheng2024navila}, a recent open-source VLN model. A heuristic viewpoint selection step is then applied to orient the camera toward the target object.

\noindent\textit{(c) VLM Prompting (VLM).}
We also include a general-purpose VLM baseline. Given visual context and language instructions, the model is prompted to output a camera location and a look-at point on the image-space.
This output representation leverages the strength of VLMs in 2D pointing. 

\paragraph{Metrics.}
\label{sec:metrics}

Evaluating language-conditioned active perception is inherently challenging due to the openness of language intents and the multi-modality of valid viewpoints.
We design the following set of complementary metrics\footnote{Unless otherwise noted, uncertainty intervals are 95\% non-parametric bootstrap intervals over test cases.}. See full details in \appref{app:metrics}.

\noindent\textit{(a) Constraint Violation (CV).}
We define the constraint violation rate as the fraction of predicted view that misses the target object or cause a collision.
This metric captures the most obvious failures.

\noindent\textit{(b) Pose Error ($e_t, e_r$).}
We report translation and rotation errors between predicted and annotated camera poses. 
But note that, these values can be misleading under multi-modal solutions, as they over-commit to a single annotated pose.


\noindent\textit{(c) Voxel-Matching F1 ($\text{F1}$).}
We further design a content-centric measure of viewpoint quality, with tolerance to minor disturbances.
A 3D voxel covered by the ground-truth view is considered \textit{matched} if there exists a voxel covered by the predicted view in its neighborhood (approximately 10 cm).
We then compute precision, recall, and the F1 score over matched voxel sets.
We regard Voxel-Matching F1 as the most reliable ``absolute'' measure.


\noindent\textit{(d) Judge Ranking (Rank).}
We further implement a list-wise ranking protocol to provide a reliable ``relative'' measure of viewpoint quality.
Given a set of candidate views generated by different methods, we use GPT-5.4 and human experts to rank the candidates, and report the average ranks and winning rates.
See detailed implementations in \appref{app:ranking}.


\subsection{Performance Comparison}
\label{sec:main_results}

\begin{figure*}[!t]
  \centering
  \includegraphics[width=0.7\textwidth]{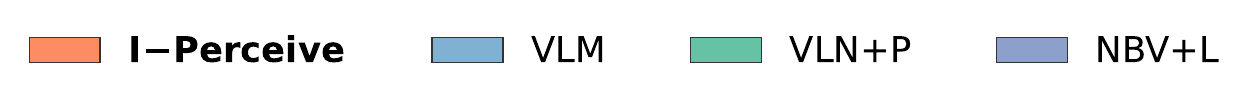}
  \vspace{2pt}

  \begin{minipage}[t]{0.19\textwidth}
    \centering
    \includegraphics[width=\linewidth]{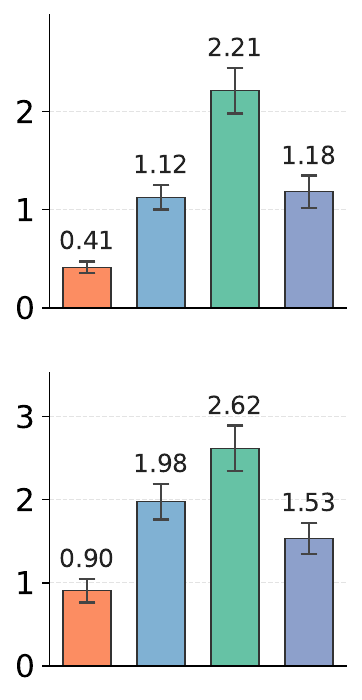}
    \caption*{\small (a) $e_t (m)\downarrow$}
  \end{minipage}
  \hfill
  \begin{minipage}[t]{0.19\textwidth}
    \centering
    \includegraphics[width=\linewidth]{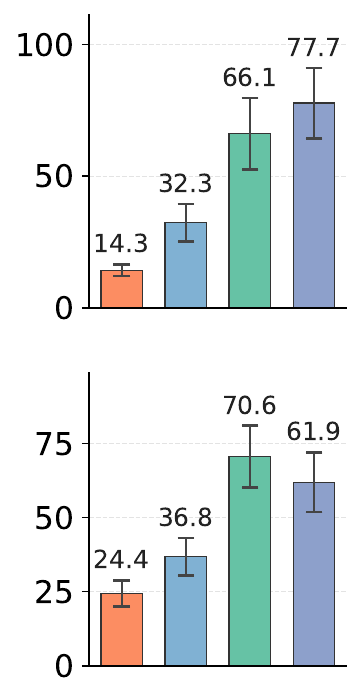}
    \caption*{\small (b) $e_r (deg) \downarrow$}
  \end{minipage}
  \hfill
  \begin{minipage}[t]{0.19\textwidth}
    \centering
    \includegraphics[width=\linewidth]{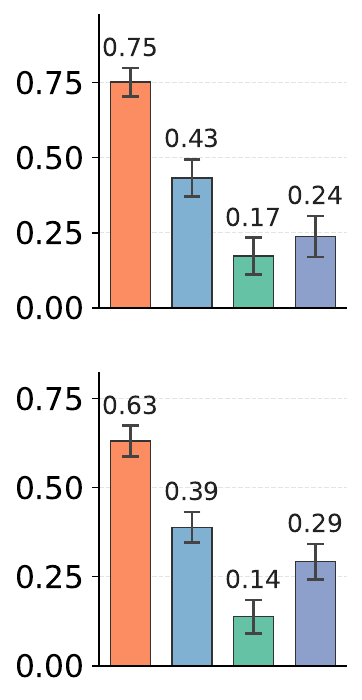}
    \caption*{\small (c) $F1 \uparrow$}
  \end{minipage}
  \hfill
  \begin{minipage}[t]{0.19\textwidth}
    \centering
    \includegraphics[width=\linewidth]{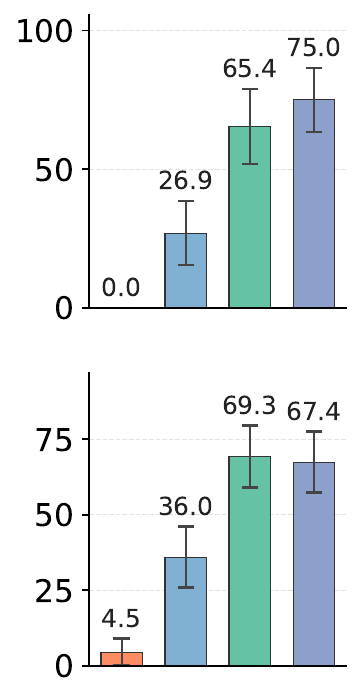}
    \caption*{\small (d) $CV (\%) \downarrow$}
  \end{minipage}
  \hfill
  \begin{minipage}[t]{0.19\textwidth}
    \centering
    \includegraphics[width=\linewidth]{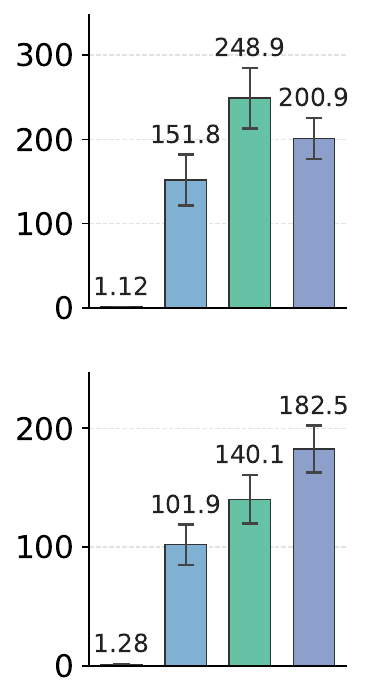}
    \caption*{\small (e) Runtime $(s) \downarrow$}
  \end{minipage}
  \caption{
  Absolute performance of \algacro and baselines on HSSD (Upper) and Replica (Bottom).
  Detailed numerical results are provided in \appref{app:main_results}.
  }
  \label{fig:main_results}
  \vspace{-10pt}
\end{figure*}

\begin{figure*}[!t]
  \centering
  \begin{minipage}[t]{0.49\textwidth}
    \centering
    \includegraphics[height=1.3cm]{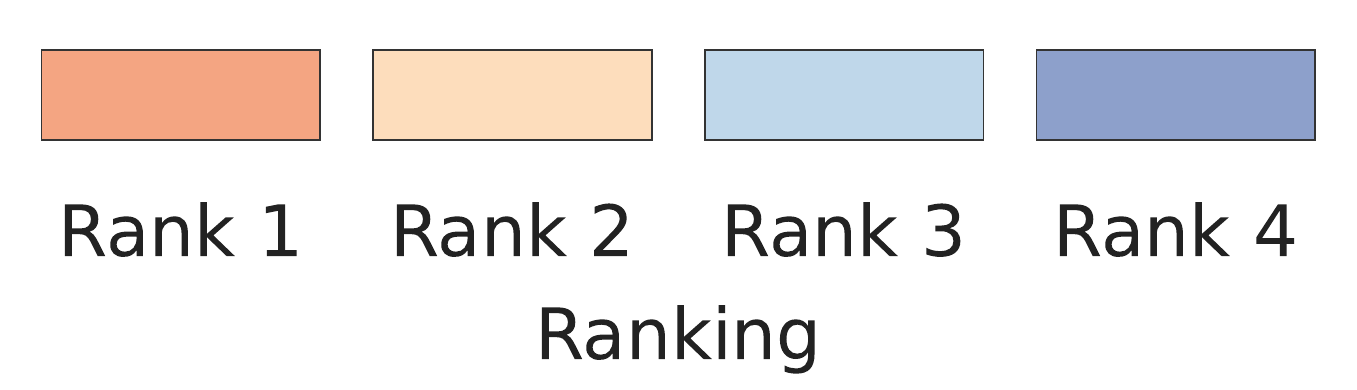}
  \end{minipage}
  \hfill
  \begin{minipage}[t]{0.49\textwidth}
    \centering
    \includegraphics[height=1.3cm]{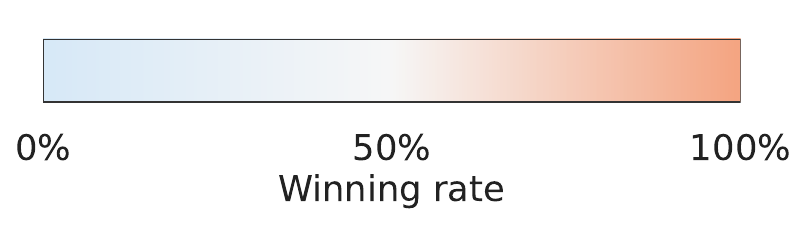}
  \end{minipage}
  \par\vspace{2pt}

  \begin{minipage}[t]{0.24\textwidth}
    \centering
    \includegraphics[width=\linewidth]{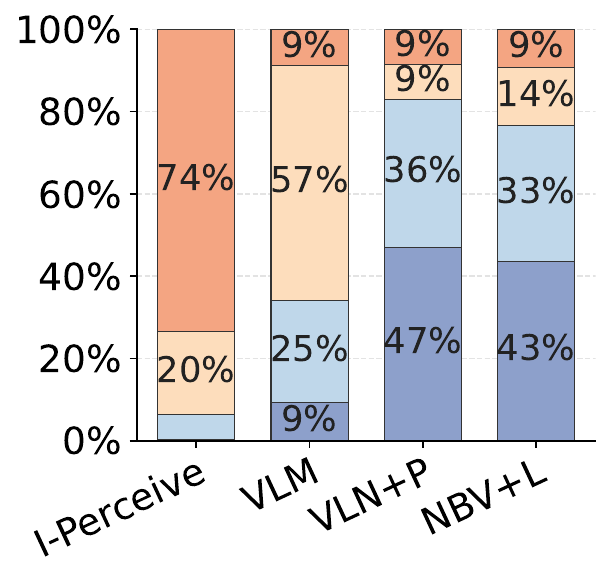}
    \caption*{\small (a) GPT ranking}
  \end{minipage}
  \hfill
  \begin{minipage}[t]{0.24\textwidth}
    \centering
    \includegraphics[width=\linewidth]{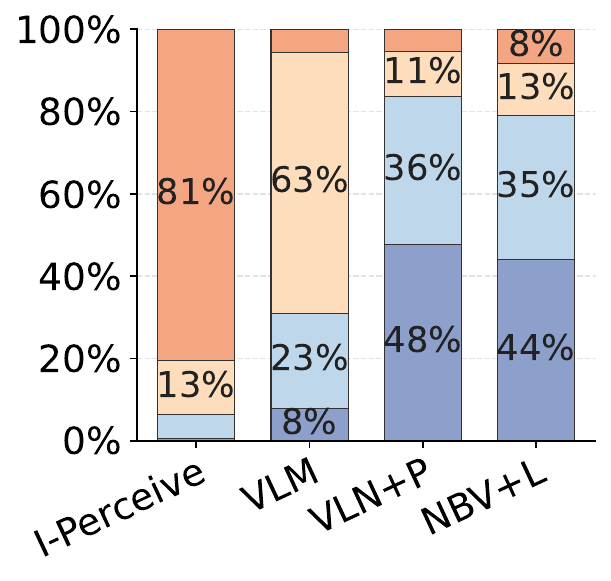}
    \caption*{\small (b) Human ranking}
  \end{minipage}
  \hfill
  \begin{minipage}[t]{0.24\textwidth}
    \centering
    \includegraphics[width=\linewidth]{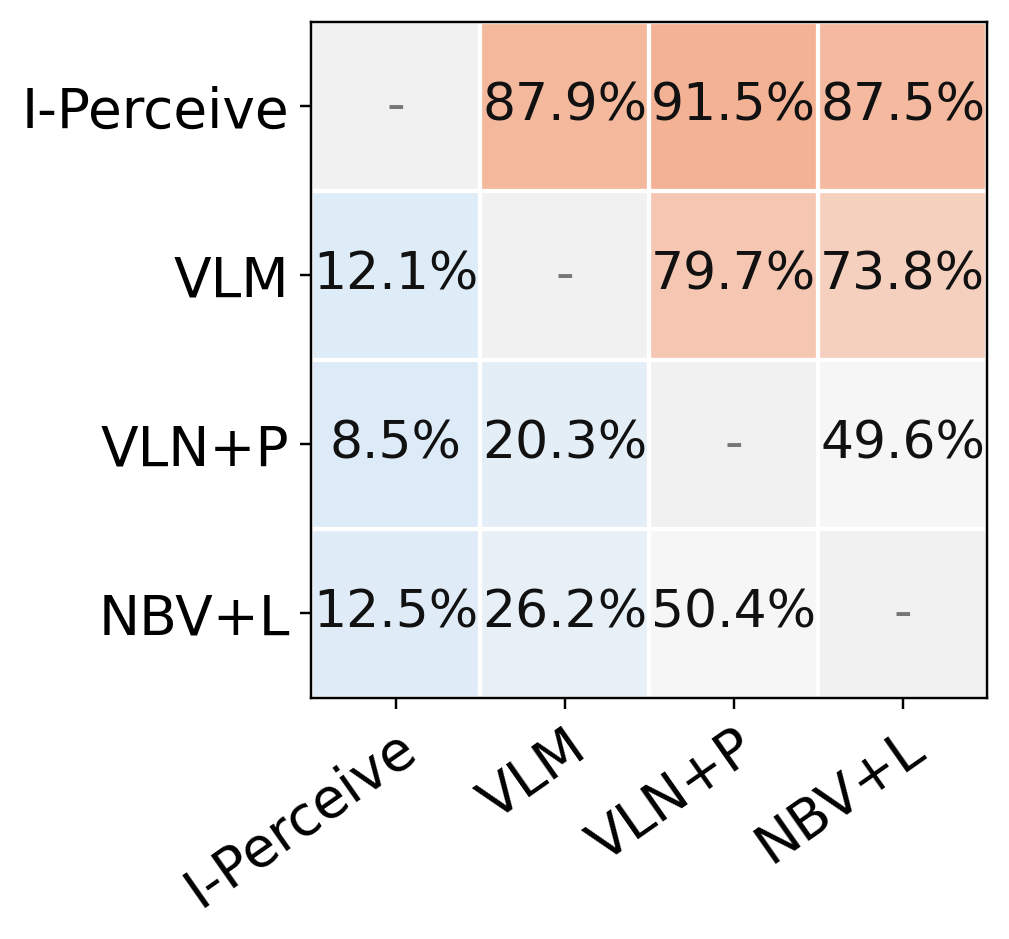}
    \caption*{\small (c) GPT winning rate}
  \end{minipage}
  \hfill
  \begin{minipage}[t]{0.24\textwidth}
    \centering
    \includegraphics[width=\linewidth]{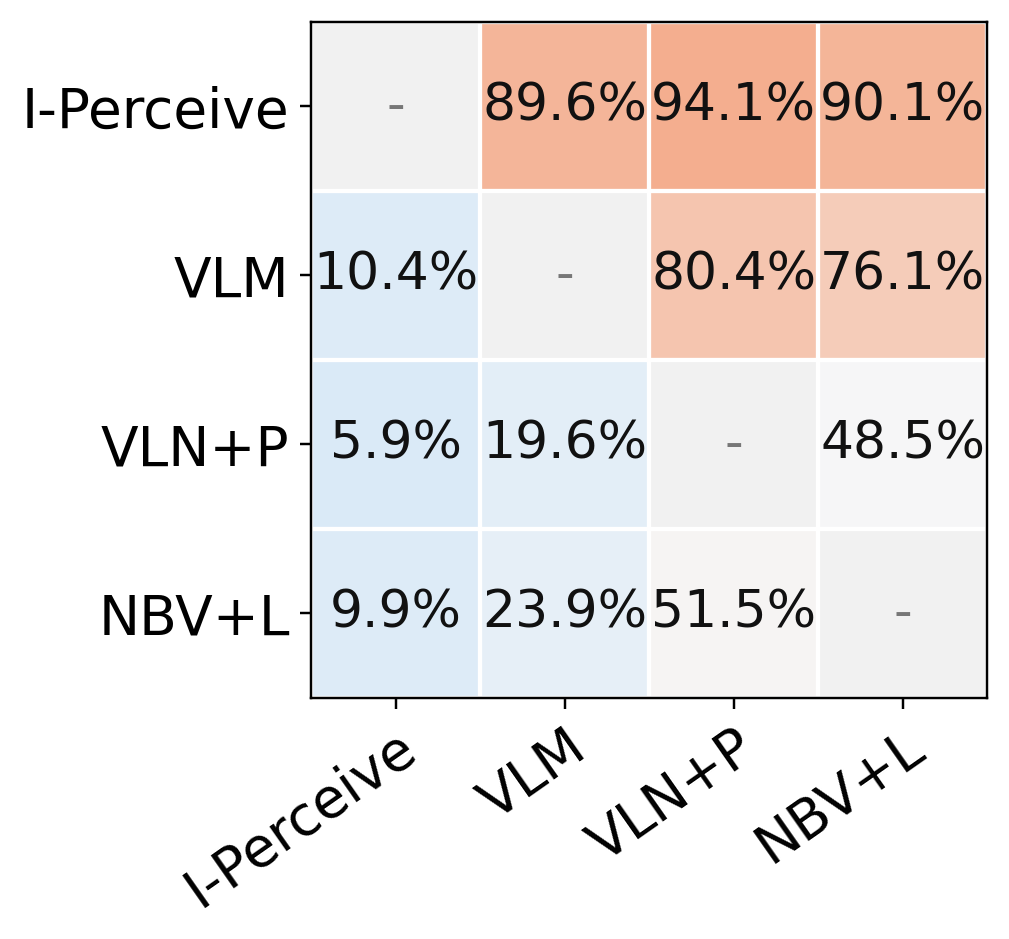}
    \caption*{\small (d) Human winning rate}
  \end{minipage}
  \caption{
  Judge ranking results of \algacro and baselines on HSSD and Replica.
  (a,b) show the distribution of ranks assigned to each method by GPT and human judges. (c,d) show the pairwise winning computed from the rankings.
  Rank 1 is best; the winning rate matrix captures row-to-column comparison.
  }
  \label{fig:rank_results}
  \vspace{-10pt}
\end{figure*}

\noindent\textbf{Absolute performance.}
\figref{fig:main_results} summarizes the performance of \algacro and all baselines on both in-distribution (HSSD) and out-of-distribution (Replica) settings.
Across both settings, \algacro consistently achieves the best performance with the lowest runtime, yielding substantially lower constraint violation, translation and rotation errors, and significantly higher voxel-matching F1 scores.
Among the baselines, 
VLN+P is the worst-performing method because its heuristic view selection disregards language intent.
NBV+L performs similarly, due to its assumption of fully observable targets.
VLM prompting yielded reasonable task performance, but significantly lower view qualities than \algacro, despite using a much longer inference time. 
In contrast, \algacro produced highly consistent views aligned with human annotations, with near-zero constraint violations, through language-conditioned geometric reasoning trained on a wide set of active perception tasks.

\noindent\textbf{Relative performance.}
\figref{fig:rank_results} reports relative performance under the judge-ranking protocol.
\algacro is ranked significantly higher than all baselines in both in-distribution and out-of-distribution settings, and achieves high pairwise winning rates ($87.5\%$--$94.1\%$) over all baselines.
Human and GPT rankings are highly consistent.

\noindent\textbf{Inference time.}
Finally, \algacro is substantially more efficient than all baselines, as it requires only a single forward pass for viewpoint prediction.
The mean inference time is $1.1$s on HSSD and $1.3$s on Replica, which involve an average of $2.6$ and $3.2$ context frames per test case, respectively.
Appendix \ref{app:ci_runtime} further provides runtime and VRAM analysis under varying context sizes.

\subsection{Performance on Close-Loop Active Perception}
\label{sec:close_loop_results}

\figref{fig:closedloop_ablation}(a) shows \algacro's performance when operating in a closed-loop setting. In this setting, no initial context is provided (single start image), and the context must be progressively collected via closed-loop execution. The targets are completely out of view at start, requiring active exploration.   
Results show that I-Perceive consistently improves performance over successive steps, in the final step, nearly matches the oracle performance with full context, demonstrating effective closed-loop exploration. 
It is worth noting that \algacro is not explicitly trained for closed-loop operation. This phenomenon emerged from \algacro's capability to accept variable-length contexts. See detailed implementations in \appref{app:closed_loop_protocol}.

\subsection{Ablation Studies}
\label{sec:ablation}

\begin{figure}[t]
\centering

\begin{minipage}[t]{0.51\linewidth}
    \centering
    \includegraphics[width=\linewidth]{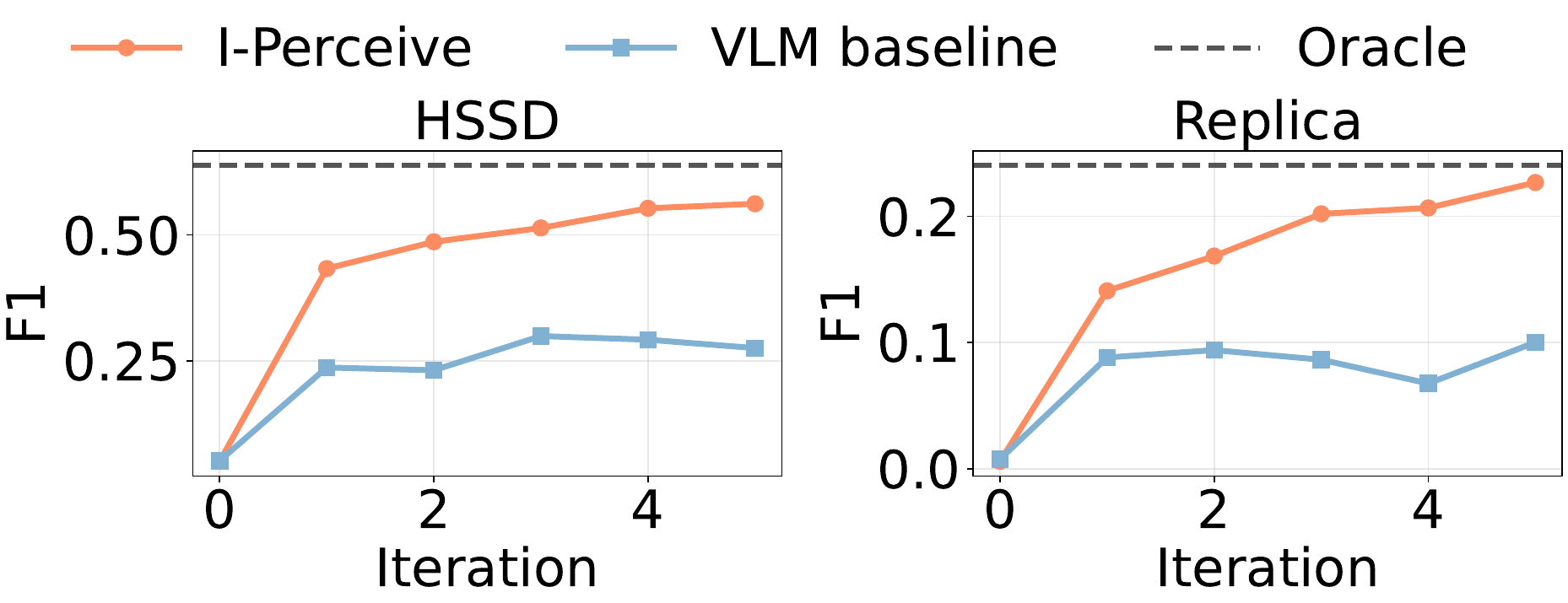}
    \caption*{{(a)}}
\end{minipage}
\hfill
\begin{minipage}[t]{0.48\linewidth}
    \centering
    \includegraphics[width=\linewidth]{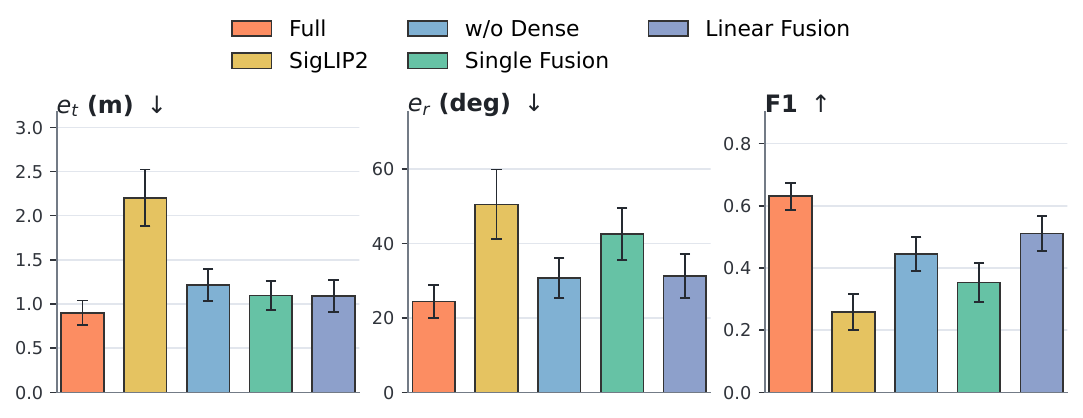}
    \caption*{{(b)}}
\end{minipage}

\caption{
Additional analysis on (a) closed-loop active perception and (b) ablation studies.
(a) Closed-loop execution of \algacro progressively improves viewpoint quality across steps.
(b) Ablation results show that replacing the VLM backbone with SigLIP2, removing dense supervision, or simplifying deep semantic fusion consistently degrades performance, highlighting their importance.
}
\label{fig:closedloop_ablation}
\vspace{-10pt}
\end{figure}

To validate key design choices, we perform ablation studies on the evaluation tasks.
Results are shown in \figref{fig:closedloop_ablation}(b). 
Replacing the VLM with a lightweight encoder (SigLIP2~\cite{tschannen2025siglip}) leads to a large performance drop, indicating that pretrained VLMs is essential for semantic understanding.
Removing auxiliary dense prediction heads (i.e., depth, point maps, and semantic masks) during training also degrades performance, due to weaker geometric and semantic reasoning.
Simplifying the fusion mechanism (i.e., fusion on a single final layer or multi-layer fusion with linear modules) harms performance, underscoring the importance of cross-attention-based progressive semantic fusion for effective language conditioning. See details of the ablated variants in \appref{app:ablation}.

\subsection{Qualitative Results}
Qualitative results, including those on in-distribution, out-of-distribution, and real-world evaluation, as well as closed-loop execution, are available through the \projecturl.


\section{Conclusion} 
\label{sec:conclusion}


We presented \algacro, a foundation model for language-conditioned active perception in large-scale indoor environments. By integrating semantic grounding from vision-language models with multi-view geometric reasoning, \algacro directly predicts task-conditioned camera poses from visual context and natural language instructions. Extensive experiments demonstrate strong performance across both in-distribution and out-of-distribution environments, significantly outperforming existing active perception, VLM prompting, and VLN-based baselines in viewpoint accuracy, constraint satisfaction, preference-based evaluation, and runtime. In addition, \algacro naturally supports closed-loop active perception, progressively improving viewpoint quality through sequential interactions under partial observability.


\noindent\textbf{Limitations and Future Work.}
Our current constraint checking only detects collisions of generated viewpoints without correcting them. Future work will incorporate collision-aware viewpoint optimization through gradient-based or sampling-based refinement. In addition, the current training pipeline remains closely tied to object annotations in scene-scanning datasets and simulation environments. Extending to richer and more open-ended targets will require stronger integration with generative world models. Finally, the dual-pathway architecture relies heavily on pretrained VLMs and geometric foundation models. While effective in limited-data regimes, future work will explore unified architectures that can better exploit large-scale training data.

\newpage
\bibliographystyle{plainnat}
\bibliography{references}

\newpage
\appendix

\newcommand{\PDataInst}{\mbox{P-DATA-INST}}
\newcommand{\PDataVideo}{\mbox{P-DATA-VIDEO}}
\newcommand{\PNBVTarget}{\mbox{P-NBVL-LOC}}
\newcommand{\PNBVAzimuth}{\mbox{P-NBVL-AZ}}
\newcommand{\PNBVElevation}{\mbox{P-NBVL-EL}}
\newcommand{\PNBVDistance}{\mbox{P-NBVL-DIST}}
\newcommand{\PNBVRefine}{\mbox{P-NBVL-REFINE}}
\newcommand{\PVLNTranslate}{\mbox{P-VLNP-NAV}}
\newcommand{\PVLNPitch}{\mbox{P-VLNP-PITCH}}
\newcommand{\PVLMCamera}{\mbox{P-VLM-POSE}}

\section{Dataset and Data Generation Details}
\label{app:data}

This appendix provides implementation-level details for the data, baselines, and evaluation protocol used in the main paper. Prompt files are referenced by compact aliases rather than copied in full; the complete prompts are included in the supplementary prompt folder. The alias mapping is shown in \tabref{tab:prompt_aliases}.

\begin{table}[h]
\centering
\caption{Prompt aliases used in this appendix.}
\label{tab:prompt_aliases}
\resizebox{\linewidth}{!}{
\begin{tabular}{lll}
\csname toprule\endcsname
Alias & File & Role \\
\midrule
\PDataInst{} & \texttt{prompts/data\_instruction\_generation.txt} & Generates short active-perception instructions. \\
\PDataVideo{} & \texttt{prompts/data\_video\_motion\_translation.txt} & Translates instructions into first-person video descriptions. \\
\PNBVTarget{} & \texttt{prompts/baseline\_nbvl\_target\_localization.txt} & Localizes the target frame and 2D box for NBV+L. \\
\PNBVAzimuth{} & \texttt{prompts/baseline\_nbvl\_azimuth\_selection.txt} & Selects NBV+L azimuth candidates. \\
\PNBVElevation{} & \texttt{prompts/baseline\_nbvl\_elevation\_selection.txt} & Selects NBV+L elevation candidates. \\
\PNBVDistance{} & \texttt{prompts/baseline\_nbvl\_distance\_selection.txt} & Selects NBV+L distance candidates. \\
\PNBVRefine{} & \texttt{prompts/baseline\_nbvl\_pose\_refinement.txt} & Applies final NBV+L pose refinement. \\
\PVLNTranslate{} & \texttt{prompts/baseline\_vlnp\_instruction\_translation.txt} & Converts active-perception requests into VLN commands. \\
\PVLNPitch{} & \texttt{prompts/baseline\_vlnp\_pitch\_refinement.txt} & Applies VLN+P pitch refinement. \\
\PVLMCamera{} & \texttt{prompts/baseline\_vlm\_camera\_lookat\_prediction.txt} & Predicts camera and look-at points for the direct VLM baseline. \\
\bottomrule
\end{tabular}
}
\end{table}

\subsection{Unified Data Format}
\label{app:data_format}

All training and evaluation samples are represented as language-conditioned viewpoint tuples. A sample contains a start frame, zero or more context frames, a target frame, and metadata:
\begin{itemize}
	\item RGB images for the start, context, and target views.
	\item Camera poses ordered as start, context frames, and target frame.
	\item Optional depth maps and semantic segmentation maps, used for auxiliary supervision and geometric evaluation.
	\item A natural-language instruction and optional 2D point reference in the start image.
	\item Camera intrinsics and scene metadata, including a scene identifier for grouped evaluation.
\end{itemize}
Each case is stored as a self-contained sample containing RGB frames, camera poses, optional dense annotations, and metadata. Camera poses follow the OpenCV camera convention: $X$ right, $Y$ down, and $Z$ forward. Unless otherwise stated, camera poses are represented as camera-to-world transformations in the corresponding scene coordinate frame, while model-predicted camera parameters are expressed relative to the start frame.

\subsection{Real-World Scanning Data}
\label{app:real_data}

For ScanNet and CA-1M, target viewpoints are sampled from existing scene-scanning trajectories. Since these datasets were not captured specifically for active perception, we use object-centric heuristics to identify useful observations. For ScanNet, semantic masks are used to select frames where a target category occupies a sufficiently visible, centered region. For CA-1M, object bounding boxes and semantic captions are used instead. Candidate target frames are filtered to remove views where the target is too small, heavily occluded, poorly centered, or geometrically inconsistent with the sampled observation intent.

Given a target viewpoint, start and context frames are sampled from nearby frames in the original scanning sequence. For the CA-1M world-model generation pipeline, the generator considers three start/context strategies: retaining the original query frame, selecting another nearby target-visible frame, or selecting nearby target-invisible frames to create harder partial-observability cases. The neighborhood window is currently set to approximately $\pm 150$ frames, corresponding to about five seconds for a 30 FPS video.

\subsection{HSSD Simulation Data}
\label{app:hssd_data}

For HSSD, we generate active-perception samples from high-quality 3D indoor scenes with controllable rendering and geometry. The HSSD pipeline starts from reusable target-pose templates attached to object categories or object instances. Each template specifies a look-at region, allowable viewing directions, distance ranges, and a short objective description. These templates support both category-level annotation for geometrically regular object classes and instance-level annotation for object categories with large shape variation.

For each scene, collision-free paths are planned between sampled target poses. Start and context frames are then sampled along these trajectories to mimic how a mobile robot accumulates observations over time. RGB images are rendered photorealistically, while depth and semantic segmentation maps are rendered from the simulator geometry. Human annotators use an interactive 3D interface to place target views, inspect start/context/target observations, and verify target visibility for benchmark cases.

\begin{figure}[t]
	\centering
    \begin{subfigure}[b]{0.24\linewidth}
        \centering
        \includegraphics[height=2.8cm, width=\linewidth]{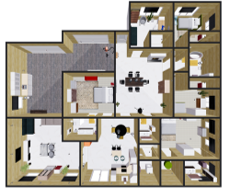} 
        \caption{HSSD Scene}
    \end{subfigure}
    \hfill
    \begin{subfigure}[b]{0.24\linewidth}
        \centering
        \includegraphics[height=2.8cm, width=\linewidth]{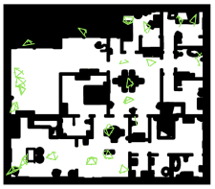}
        \caption{Annotated target frame}
    \end{subfigure}
    \hfill
    \begin{subfigure}[b]{0.24\linewidth}
        \centering
        \includegraphics[height=2.8cm, width=\linewidth]{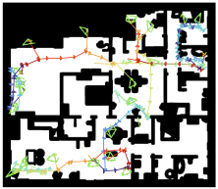}
        \caption{Generated trajectories}
    \end{subfigure}
    \hfill
    \begin{subfigure}[b]{0.24\linewidth}
        \centering
        \includegraphics[height=2.8cm, width=\linewidth]{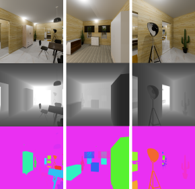}
        \caption{Rendered frame}
    \end{subfigure}
	\caption{\textbf{Data generation pipeline in simulated HSSD environments.} (a) Top-down view of a representative 3D house scene. (b) Target viewpoints (green frustums) are uniformly sampled across the navigable floor map. (c) To mimic real-world online exploration, robot movement trajectories towards the target viewpoints are generated using path planning. (d) Context frames are sampled along the movement history, providing dense multi-modal learning signals including RGB, depth maps, and semantic masks.}
	\label{fig:app_hssd_annotation_placeholder}
\end{figure}

\subsection{Instruction Generation and 2D Point Disambiguation}
\label{app:instruction_generation}

Natural-language instructions are generated by a VLM conditioned on the target view and available semantic information. For ScanNet, the semantic category label is provided. For CA-1M, the object caption and bounding box are provided. For HSSD, the prompt can additionally use target-pose objective descriptions and rendered start/target images.

The instruction generator is constrained to produce short, action-oriented active-perception commands that describe where the agent should look or move its camera, rather than long analytical goals. It also avoids ego-centric references such as ``turn left'' because the same instruction may be paired with different start frames during training. The relevant prompt aliases are:
\begin{itemize}
	\item \PDataInst{}: active-perception instruction generation.
	\item \PDataVideo{}: translation from an active-perception instruction to a first-person video description for the world model.
\end{itemize}
When an instruction is ambiguous because multiple objects or regions could satisfy it, the data pipeline can add a 2D point reference in the start image. The text instruction and optional point are represented with a compact JSON-style schema so the model observes a consistent instruction format.

\subsection{World-Model Generated Post-Training Data}
\label{app:world_model_data}

The enhanced post-training data is generated with PerceiveBench. The pipeline first samples a key frame from CA-1M whose center points to a meaningful object. A VLM receives the image, the target bounding box, and the semantic caption, and generates a short active-perception instruction. A second prompt translates the instruction into a first-person video description. Wan2.2 then synthesizes a static-scene, camera-motion video, and VGGT estimates the camera motion from sampled video frames. The final frame pose is used as an improved target pose for post-training.

The generation pipeline is run as an offline data-production workflow using remote VLM, video-generation, and pose-estimation services. Generated CA-1M samples are validated before use; invalid samples are separated by failure reason, including malformed model outputs, failed pose recovery, and geometric validation failures.

\subsection{Manual annotation in evaluation set}
\label{app:manual_annotation}

To ensure the high fidelity and semantic consistency of our evaluation suite, we conducted a rigorous manual annotation process for the HSSD and Replica datasets. All annotations were performed by researchers specialized in computer vision and robotics to guarantee professional-grade data quality.

\begin{figure}[htbp]
    \centering
    \includegraphics[width=0.9\linewidth]{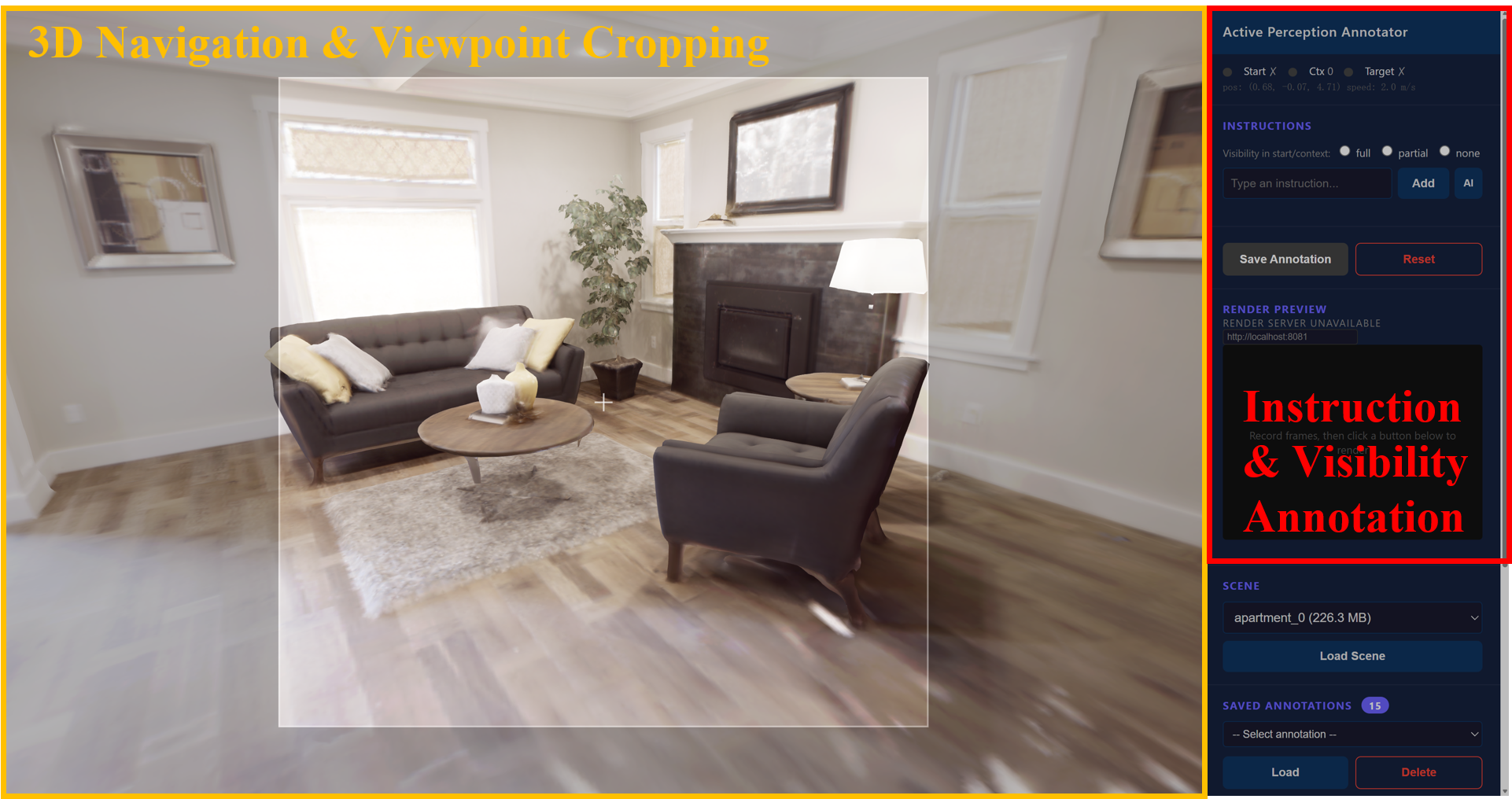} 
    \caption{\textbf{Active Perception Annotation Interface.} The tool is designed to ensure rigorous evaluation set construction. 
    \textbf{(A)} The main viewport allows annotators to navigate 3D environments and determine the optimal field of view (highlighted central square). 
    \textbf{(B)} The keyframe tracker records the precise 6D poses of the \textit{start}, \textit{context}, and \textit{target} views. And the instruction module where annotators provide natural language guidance and assess object visibility. }
    \label{fig:annotation_ui}
\end{figure}

\paragraph{Task Definition}

For each data point, annotators were tasked with identifying three key viewpoints: the start view, context view, and target view. Specifically, given a scene, annotators had to:

\begin{itemize}
    \item Identify a coherent sequence of views representing the task's progression.
    \item Formulate a natural language instruction that accurately describes the transition to the target state.
    \item Verify that the selected target view represents the optimal viewpoint for fulfilling the given instruction within that scene.
\end{itemize}

\paragraph{Training and Standardization}

Before the formal annotation phase, all annotators underwent a unified training session. This training focused on standardizing the subjective understanding of the "optimal viewpoint" and "active perception." We established clear criteria for viewpoint quality, such as occlusion-free visibility of target objects and appropriate camera-to-object distances, ensuring that the subjective bias across different annotators was minimized.

\paragraph{Quality Control}

We implemented a multi-stage quality assurance pipeline:

\begin{enumerate}
    \item Self-Inspection: Annotators were required to re-evaluate each sequence immediately after labeling to ensure the instruction-view alignment.
    \item Secondary Verification: Upon completion, a secondary screening was performed by a separate expert. Any sequences with ambiguous instructions or suboptimal target poses were either refined or discarded. This rigorous filtering ensures that our evaluation set serves as a gold standard for assessing active perception capabilities.
\end{enumerate}

\begin{figure}[t]
	\centering
    \begin{subfigure}[b]{0.24\linewidth}
        \centering
        \includegraphics[height=2cm, width=\linewidth]{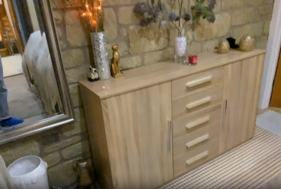} 
        \caption{Input frame}
    \end{subfigure}
    \hfill
    \begin{subfigure}[b]{0.24\linewidth}
        \centering
        \includegraphics[height=2cm, width=\linewidth]{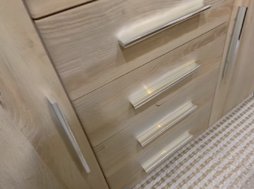}
        \caption{Video's endframe}
        \label{fig:video_endpoint}
    \end{subfigure}
    \hfill
    \begin{subfigure}[b]{0.24\linewidth}
        \centering
        \includegraphics[height=2cm, width=\linewidth]{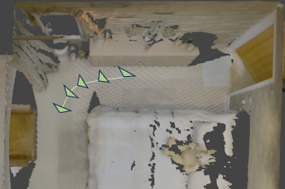}
        \caption{Recovered target pose}
        \label{fig:target_pose}
    \end{subfigure}
    \hfill
    \begin{subfigure}[b]{0.24\linewidth}
        \centering
        \includegraphics[height=2cm, width=\linewidth]{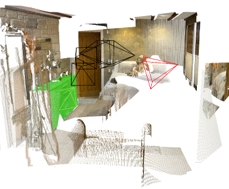}
        \caption{Resampled contexts}
        \label{fig:additional}
    \end{subfigure}
	\caption{ \textbf{The world model data generation pipeline.} (a) The initial frame from the scanning dataset. (b) The endframe of the video synthesized by the world model, which serves as the new target view. (c) The recovered camera trajectory and target pose estimated via VGGT. (d) 3D scene reconstruction and context frame resampling to formulate the final training data point.}
	\label{fig:world_model_data_generation_pipiline}
\end{figure}

\section{Model and Training Details}
\label{app:model_training}

\subsection{Backbones and Fusion Layers}
\label{app:model_details}

The vision-language pathway uses Qwen3-VL-4B to encode the image sequence and instruction. Rather than using only the final hidden state, we extract intermediate key-value features from selected transformer layers and inject them into the geometric reasoning pathway. We collect VLM KV features from layers $[8,15,23,35]$ in zero-based indexing and fuse them into geometric layers $[0,5,12,18]$. In one-based notation, this corresponds to VLM layers $[9,16,24,36]$ and geometric layers $[1,6,13,19]$.

The geometric pathway is based on a VGGT-style multi-view transformer. Each frame contains a camera token, register tokens, semantic tokens, and patch tokens. Semantic tokens are placed on the same patch grid as visual tokens and serve as carriers for language-grounded information. For observed image frames, semantic tokens attend to VLM image KV features with 2D rotary position embeddings. For the latent instruction/target frame, semantic tokens attend to instruction KV features with 1D rotary position embeddings on the keys. The cross-attention update is gated by a learnable LayerScale parameter initialized to zero for stable training.

\subsection{Prediction Heads and Losses}
\label{app:losses}

The camera head predicts the target camera parameters and the observed-frame camera parameters. Auxiliary heads produce depth maps, 3D point maps, 2D grounding heatmaps with visibility logits, and optional target-object semantic masks. During inference, only the camera prediction is required. During training, the total objective is a weighted sum of the following components:
\begin{itemize}
	\item camera pose loss for translation, rotation, and focal representation;
	\item depth loss and confidence loss;
	\item 3D point-map loss and confidence loss;
	\item grounding heatmap, visibility, and soft-argmax coordinate losses;
	\item semantic-mask loss combining pixel-wise and region-level supervision.
\end{itemize}

\subsection{Training Schedule}
\label{app:training_schedule}

We first train on the pre-training set for 5 epochs with the VLM backbone frozen and all learning targets enabled. We then fine-tune on the enhanced post-training set for 3 epochs, using LoRA to adapt the VLM attention projections and disabling the semantic mask and point-heatmap losses to emphasize instruction following. The model is trained with mixed precision: parameters are kept in fp32 during training, forward computation uses bf16 autocast by default, and released inference checkpoints are stored in bf16.

\section{Baseline Designs}
\label{app:baselines}

All methods are evaluated through the same pose-evaluation protocol. Each method receives the same start/context images, camera metadata, and instruction, and returns a single predicted camera-to-world pose. The returned pose is rendered and scored with the same metric definitions across methods.

\subsection{NBV+L: Language-Conditioned Next-Best-View}
\label{app:baseline_nbvl}

NBV+L is a training-free, VLM-guided next-best-view baseline. It first asks a VLM to select the relevant observed frame and target bounding box. A frozen VGGT-1B model estimates geometry from the start and context frames; the prediction is refined by aligning the start-frame depth scale to the known data depth. The selected 2D target is back-projected to 3D, and a sparse occupancy model is built for collision and line-of-sight checks.

After target localization, the baseline decomposes view selection into interpretable low-cardinality decisions: azimuth selection over eight compass directions, elevation selection over five vertical angles, distance selection over the metric grid $\{0.3,0.8,1.3,1.8,2.3,2.8\}$ meters, and a final semantic micro-refinement over yaw, pitch, and distance. The corresponding prompt aliases are \PNBVTarget{}, \PNBVAzimuth{}, \PNBVElevation{}, \PNBVDistance{}, and \PNBVRefine{}.

\subsection{VLN+P: Vision-Language Navigation with Perception Refinement}
\label{app:baseline_vlnp}

VLN+P reformulates active perception as a Vision-Language Navigation (VLN) problem. NaVILA is a VLM-based VLN model that maps visual observations and navigation instructions to discrete movement actions. A VLM first translates the active-perception instruction into a NaVILA-style navigation command using \PVLNTranslate{}. NaVILA then performs an online rollout from the start pose, using rendered observations at each step. Actions are parsed into \texttt{stop}, forward moves at 25 cm granularity, or left/right turns at 15 degree granularity. The rollout stops when NaVILA predicts stop, after at most 20 steps, or after repeated collision-blocked forward moves. Since VLN policies target goal-reaching rather than viewpoint selection, a final VLM call applies a pitch correction clamped to $[-45^\circ,45^\circ]$ using \PVLNPitch{}.

\subsection{VLM Prompting}
\label{app:baseline_vlm}

The VLM baseline removes VGGT geometry, candidate sampling, and occupancy checks. Given the start image, context images, instruction, and optional target point hint, the VLM emits two 3D marks in the start-frame coordinate system: a camera point and a look-at point. Each mark is represented as a per-mille image coordinate $(u,v)$ and a metric depth. These points are back-projected through the known start-frame intrinsics and pose, and the final camera is constructed with a gravity-up look-at transform. Degenerate predictions where the two points nearly coincide are repaired by nudging the look-at point forward along the start camera's $+Z$ axis. The prompt alias is \PVLMCamera{}.

\subsection{Human Evaluation for Model Predictions (Rank-H)}
\label{sec:rank_h_interface}

To conduct a rigorous human evaluation of the generated target views, we developed a dedicated listwise ranking interface, as shown in \textbf{Figure \ref{fig:ranking_ui}}. Consistent with the evaluation set construction process, all Rank-H judgments were performed by the same group of domain experts to ensure high-quality and reliable assessments.

\paragraph{Interface Design and Anonymization}
To prevent evaluation bias, all model predictions are strictly anonymized and their spatial positions (A, B, C, D) are randomly shuffled for each test instance. As illustrated in the interface, the evaluator is provided with comprehensive context:
\begin{itemize}
    \item \textbf{Task Instruction:} The natural language command defining the semantic goal (e.g., ``Prepare to open the oven'').
    \item \textbf{Reference Sequence:} The complete trajectory including the start view, intermediate context views, and the ground-truth (GT) target view.
    \item \textbf{Anonymized Predictions:} Four candidate target views comprising the output of our proposed method and three established baselines.
\end{itemize}

\paragraph{Ranking Protocol and Criteria}
Evaluators are tasked with providing a strict, forced-choice ranking of the four candidates from 1 (best) to 4 (worst), with \textbf{no ties permitted}. The primary evaluation criterion is the semantic alignment between the generated view and the text instruction, ensuring the output is visually consistent with the underlying 3D scene geometry. Crucially, the GT target view is provided merely as a contextual reference to help evaluators understand the spatial progression of the scene; it does not serve as the absolute ranking standard. Evaluators are explicitly instructed to reward models that successfully fulfill the text instruction, even if the predicted viewpoint differs from the GT. To ensure efficiency during large-scale evaluation, the interface supports seamless keyboard shortcuts for ranking assignment and navigation.

\section{Evaluation Metrics and Ranking Protocol}
\label{app:metrics}

\subsection{Pose Errors}
\label{app:pose_errors}

For each test case, we compare the predicted pose $\mathbf{g}_{\text{pred}}$ to the human-annotated reference pose $\mathbf{g}_{\text{gt}}$. The translation error is the Euclidean distance between camera centers. The rotation error is the geodesic distance on $SO(3)$, reported in degrees in the main table.

\subsection{Voxel-Matching F1}
\label{app:f1}

Let $\mathcal{V}(\mathbf{g})$ denote the set of scene voxels visible from camera pose $\mathbf{g}$ under frustum and occlusion testing. A ground-truth voxel $v\in\mathcal{V}(\mathbf{g}_{\text{gt}})$ is considered recalled if there exists a predicted voxel $u\in\mathcal{V}(\mathbf{g}_{\text{pred}})$ satisfying $\|u-v\|_{\infty}\le k$. Precision is defined symmetrically. We use $k=2$ voxel units in the main experiments, corresponding to approximately 10 cm under the 5 cm voxel pitch used for HSSD:
\begin{equation}
    \operatorname{sR}_k = \frac{|\{v \in \mathcal{V}_{\mathrm{gt}} : \exists u \in \mathcal{V}_{\mathrm{pred}}, \|u-v\|_\infty \le k\}|}{|\mathcal{V}_{\mathrm{gt}}|}, \quad
    \operatorname{sP}_k = \frac{|\{u \in \mathcal{V}_{\mathrm{pred}} : \exists v \in \mathcal{V}_{\mathrm{gt}}, \|u-v\|_\infty \le k\}|}{|\mathcal{V}_{\mathrm{pred}}|},
\end{equation}
\begin{equation}
    \operatorname{F1}_k = \frac{2\,\operatorname{sP}_k\,\operatorname{sR}_k}{\operatorname{sP}_k+\operatorname{sR}_k}.
\end{equation}

\subsection{Constraint Violation}
\label{app:cv}

Constraint violation (CV) measures the fraction of test cases where the predicted view either misses the target or violates physical feasibility. We track two flags: \texttt{target\_visible}, annotated by human inspection of the rendered prediction, and \texttt{in\_collision}, computed by a sphere-collision check around the predicted camera center against the scene voxel grid. A case counts as a violation if the target is not visible or the predicted camera is in collision. Only cases with both annotations available are included in the CV denominator.




\subsection{List-Wise Judge Ranking Evaluation Details}
\label{app:ranking}

To comprehensively evaluate the generated target views, we employ a list-wise ranking protocol. For a given test case, the judge is presented with the task instruction, the reference context, and the anonymized rendered outputs of all competing methods. The judges rank candidate views according to how well they satisfy the semantic instruction and provide a physically plausible, useful observation of the target. We report the \textit{mean rank}, where a lower value indicates better performance.

\paragraph{Evaluation Groups and Aggregation}
The main results reported in our experiments are based on two distinct judge groups: an automated LLM judge, GPT-5.4 (Rank-G), and human annotators (Rank-H). To mitigate position bias, the spatial ordering (A, B, C, D) of the anonymized candidates is randomly shuffled. For each evaluation group, three independent shuffled runs are aggregated. We first average the ranks for the same test case across the three runs, and then compute the dataset-level means along with their bootstrap confidence intervals.

\begin{figure}[htbp]
    \centering
    \includegraphics[width=\linewidth]{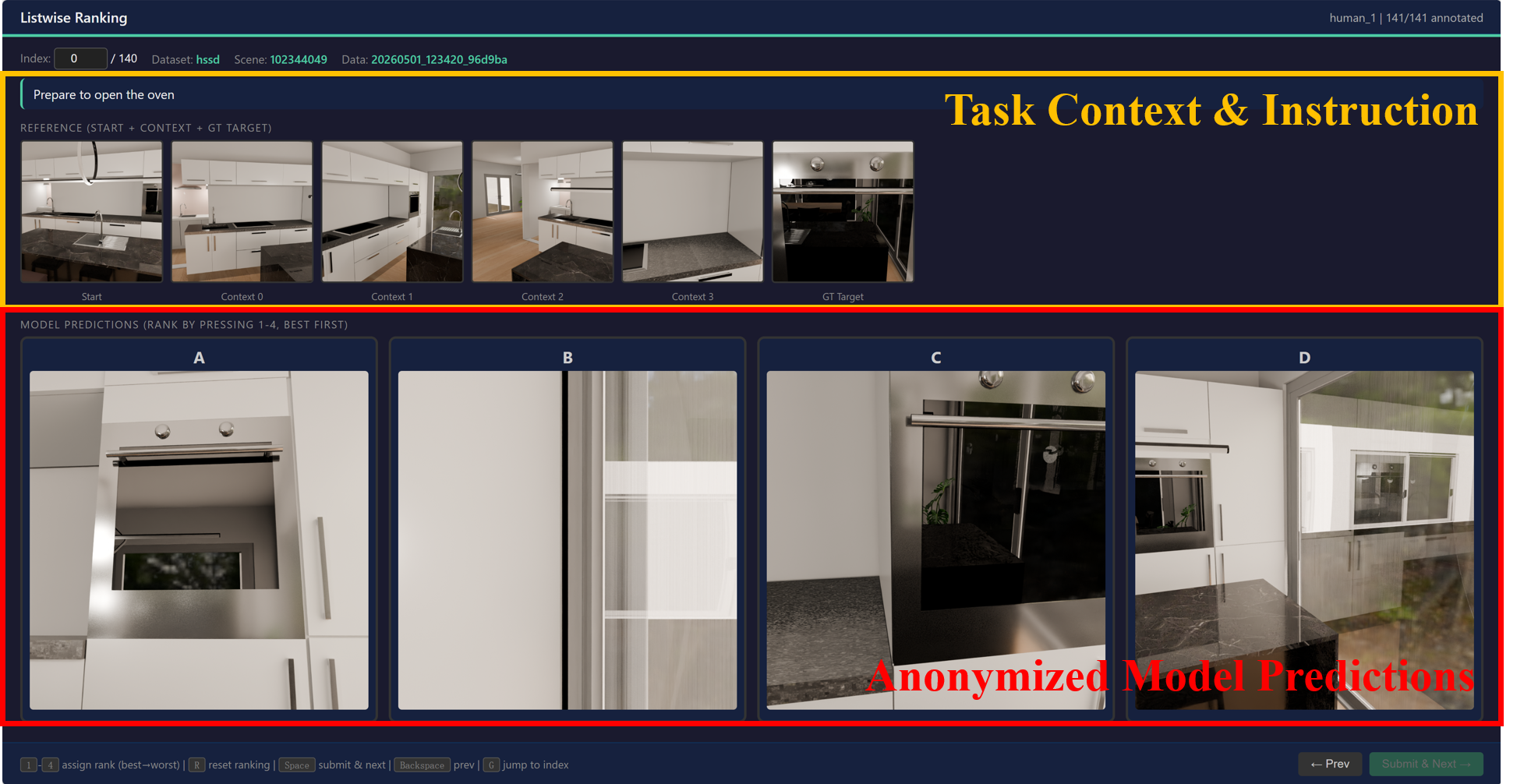} 
    \caption{\textbf{The Listwise Ranking Interface for Human Evaluation (Rank-H).} \textbf{(A)} Evaluators are provided with the natural language instruction and the reference temporal sequence (from start to ground-truth target) to understand the task context. \textbf{(B)} Four anonymized and randomly shuffled model predictions are presented for side-by-side comparison.}
    \label{fig:ranking_ui}
\end{figure}

\paragraph{Human Evaluation Interface (Rank-H)}
To conduct the human evaluation rigorously, we developed a dedicated ranking interface, illustrated in \textbf{Figure \ref{fig:ranking_ui}}. Consistent with the evaluation set construction process, all Rank-H judgments were performed by domain experts. As shown in the interface, the ground-truth (GT) target view is provided merely as a contextual reference (alongside the start and context views) to illustrate the spatial progression of the scene; it is \textbf{not} shown as a candidate to be ranked. 

The interface enforces a strict, forced-choice ranking from 1 (best) to 4 (worst) with \textbf{no ties permitted}. Evaluators are explicitly instructed to prioritize the semantic alignment with the text instruction, rewarding models that successfully fulfill the task even if the predicted viewpoint differs from the GT reference.

\subsection{Confidence Intervals and Runtime}
\label{app:ci_runtime}

For $e_t$, $e_r$, F1, Rank-G, and Rank-H, confidence intervals are computed with a non-parametric percentile bootstrap over test cases using 10,000 resamples. The table reports the mean and a conservative half-width of the 95\% interval. Runtime is reported as the mean runtime returned by each algorithm implementation and is not assigned a confidence interval. For \algacro, runtime includes model forward and pose-refinement post-processing, but excludes data adapter preprocessing and host-device transfer.

\begin{table}[!t]
	\vspace{-0.2em}
	\centering
	\caption{Runtime and VRAM usage of \algacro as a function of the number of input images on GTX 4090.}
	\label{tab:runtime}
	\small
	\begin{tabular}{l|ccccc}
		\hline
		\#Images & 1 & 2 & 3 & 4 & 5 \\
		\hline
		Runtime (s) & 0.78 & 1.42 & 2.13 & 2.90 & 3.82 \\
		VRAM (GB) & 16.07 & 16.92 & 17.84 & 18.79 & 19.74 \\
		\hline
	\end{tabular}
	\vspace{-1.2em}
\end{table}

\section{Experimental Details}
\label{app:exp_details}

\subsection{Benchmarks}
\label{app:benchmarks}

HSSD is used as the in-distribution benchmark. The HSSD test set contains 52 language-conditioned tasks from held-out 5 scenes. Target viewpoints are manually annotated by human experts rather than generated by the training-data pipeline.

Replica is used as the out-of-distribution benchmark. It contains 89 language-conditioned tasks in reconstructed real-world indoor environments.

\subsection{Evaluation Pipeline}
\label{app:evaluation_pipeline}

All quantitative methods are evaluated by a shared pipeline. The evaluator enumerates test cases, groups cases by scene to reuse voxel grids, calls the method-specific predictor, renders the predicted view through the corresponding HSSD or Replica renderer, computes metrics, and writes per-case and aggregate summaries.

\subsection{Result Analysis}
\label{app:analysis}

The main results are analyzed from per-case evaluation records. The analysis aggregates all four methods and both datasets, computes bootstrap intervals, and exports the statistics used to format the main result table.

\section{Detailed Main Results}
See Table \ref{tab:main_results}.
\label{app:main_results}

\begin{table*}[!t]
\centering
\caption{
Performance comparison on HSSD and Replica.
We report constraint violation (CV), voxel-matching F1 (F1), pose error ($e_t$, $e_r$), mean judge rank from GPT-5.4 (Rank-G) and human annotators (Rank-H), and mean runtime.
$\uparrow$: higher is better; $\downarrow$: lower is better.
}
\label{tab:main_results}
\resizebox{\textwidth}{!}{
\begin{tabular}{ll|ccccccc}
\toprule
\textbf{Dataset}
& \textbf{Method}
& $\mathbf{e_t}$ (m) $\downarrow$
& $\mathbf{e_r}$ (deg) $\downarrow$
& \textbf{F1} $\uparrow$
& \textbf{Rank-G} $\downarrow$
& \textbf{Rank-H} $\downarrow$
& \textbf{CV} (\%) $\downarrow$
& \textbf{Time} (s) $\downarrow$ \\
\midrule
HSSD & NBV+L          & $1.18 \pm 0.16$ & $77.70 \pm 13.39$ & $0.2373 \pm 0.0679$ & $3.28 \pm 0.25$ & $3.38 \pm 0.22$ & $75.0 \pm 11.5$ & $200.9 \pm 24.4$ \\
HSSD & VLN+P          & $2.21 \pm 0.23$ & $66.06 \pm 13.59$ & $0.1726 \pm 0.0610$ & $3.21 \pm 0.21$ & $3.23 \pm 0.19$ & $65.4 \pm 13.5$ & $248.9 \pm 35.9$ \\
HSSD & VLM            & $1.12 \pm 0.12$ & $32.33 \pm 7.15$  & $0.4322 \pm 0.0617$ & $2.25 \pm 0.19$ & $2.21 \pm 0.16$ & $26.9 \pm 11.5$ & $151.8 \pm 30.2$ \\
HSSD & \textbf{\algacro} & $\mathbf{0.41 \pm 0.06}$ & $\mathbf{14.31 \pm 2.25}$ & $\mathbf{0.7511 \pm 0.0474}$ & $\mathbf{1.26 \pm 0.15}$ & $\mathbf{1.18 \pm 0.14}$ & $\mathbf{0.0 \pm 0.0}$ & $\mathbf{1.1 \pm 0.1}$ \\
\midrule
Replica & NBV+L          & $1.53 \pm 0.18$ & $61.92 \pm 10.10$ & $0.2917 \pm 0.0499$ & $3.01 \pm 0.20$ & $3.01 \pm 0.19$ & $67.4 \pm 10.1$ & $182.5 \pm 19.8$ \\
Replica & VLN+P          & $2.62 \pm 0.27$ & $70.61 \pm 10.45$ & $0.1374 \pm 0.0466$ & $3.22 \pm 0.21$ & $3.28 \pm 0.19$ & $69.3 \pm 10.2$ & $140.1 \pm 20.4$ \\
Replica & VLM            & $1.98 \pm 0.21$ & $36.77 \pm 6.35$  & $0.3883 \pm 0.0431$ & $2.40 \pm 0.16$ & $2.40 \pm 0.14$ & $36.0 \pm 10.1$ & $101.9 \pm 17.0$ \\
Replica & \textbf{\algacro} & $\mathbf{0.90 \pm 0.14}$ & $\mathbf{24.41 \pm 4.40}$ & $\mathbf{0.6306 \pm 0.0430}$ & $\mathbf{1.37 \pm 0.12}$ & $\mathbf{1.31 \pm 0.13}$ & $\mathbf{4.5 \pm 4.5}$ & $\mathbf{1.3 \pm 0.0}$ \\
\bottomrule
\end{tabular}
}
\vspace{-10pt}
\end{table*}

\section{Ablation Details}
\label{app:ablation}

The ablation study compares the full reference model against four variants: replacing the VLM with SigLIP2, removing dense auxiliary tasks, using only a single fusion layer, and replacing spatial cross-attention fusion with linear fusion. All variants use the complete training recipe and are evaluated with the same metric protocol.

\begin{table*}[t]
\centering
\caption{Ablation results on the Replica dataset using the complete training recipe. $\pm$ denotes the 95\% bootstrap half-width. The full-model F1 value is rounded from $0.6306$ in \tabref{tab:main_results} to three decimals.}
\label{tab:ablation_replica}
\resizebox{\textwidth}{!}{
\begin{tabular}{l|ccc}
\toprule
Method & $e_t$ (m) $\downarrow$ & $e_r$ (deg) $\downarrow$ & F1 $\uparrow$ \\
\midrule
\textbf{\algacro full} & $\mathbf{0.90 \pm 0.14}$ & $\mathbf{24.41 \pm 4.40}$ & $\mathbf{0.631 \pm 0.043}$ \\
Replace VLM with SigLIP2 & $2.202 \pm 0.320$ & $50.48 \pm 9.34$ & $0.259 \pm 0.058$ \\
w/o Dense Tasks & $1.214 \pm 0.179$ & $30.71 \pm 5.46$ & $0.445 \pm 0.054$ \\
Single Fusion Layer & $1.096 \pm 0.165$ & $42.53 \pm 7.04$ & $0.353 \pm 0.063$ \\
Linear Fusion & $1.091 \pm 0.178$ & $31.21 \pm 5.98$ & $0.511 \pm 0.055$ \\
\bottomrule
\end{tabular}
}
\end{table*}

On Replica, the full model is clearly better across all reported metrics. Replacing the VLM with SigLIP2 causes the largest overall degradation, while removing dense auxiliary supervision and simplifying the fusion mechanism also reduce performance.

\section{Additional Qualitative and Closed-Loop Results}
\label{app:qualitative}

\subsection{Closed-Loop Evaluation Protocol}
\label{app:closed_loop_protocol}

For closed-loop evaluation, we adapt each test case to a harder partial-observability setting by removing all initial context frames. The model therefore starts from a single image, and iteration 0 reports the coverage of this initial view before any model action. At each later iteration, the method predicts a target camera pose from the current start image, instruction, and accumulated context views.

To emulate execution on a robot, the predicted pose is passed through the same lightweight execution wrapper for all compared methods, including \algacro and the VLM baseline. The wrapper applies collision-avoidance correction and rule-based exploration before rendering or collecting the next observation, and the resulting view is appended to the context for the following iteration. This design reflects a realistic deployment pipeline: a robot would not execute raw network outputs directly, but would combine them with safety checks and simple exploration heuristics before acquiring the next camera view.



\end{document}